\documentclass{article}
\usepackage{fullpage}

\usepackage[title]{appendix}

\usepackage{diagbox}
\usepackage[utf8]{inputenc} 
\usepackage[T1]{fontenc}    
\usepackage{url}            
\usepackage{multirow}
\usepackage{booktabs}       
\usepackage{amsfonts,amssymb,amsthm}       
\usepackage{nicefrac}       
\usepackage{microtype}      
\usepackage{hhline}

\newtheorem{mydef}{Definition}
\newtheorem{theorem}{Theorem}

\newtheorem{prop}{Proposition}

\newcommand{\revise}{}

\usepackage[T1]{fontenc}
\usepackage[utf8]{inputenc}
\usepackage{pgfplots}
\usepackage{graphicx}
\usepackage{subcaption}
\usepackage{pstricks}
\usepackage{color}
\usepackage{amsmath}
\usepackage{bbm}
\usepackage{tcolorbox}
\usepackage{tikz}
\usepackage{pgfplots}
\usepackage{wrapfig}
\usepackage{lipsum}  
\usetikzlibrary{positioning}
\usepackage{tcolorbox}
\usepackage{amsfonts,amssymb}
\usepackage{mathtools}
\usepackage{commath}
\usepackage{relsize}
\usepackage{bbm}
\usepackage{bm}
\usepackage[font={small}]{caption}
\usepackage{comment,color,soul}
\usepackage{array}
\newcolumntype{?}{!{\vrule width 1pt}}
\usepackage{enumerate} 
\usepackage{amsfonts}
\usepackage{url}
\usepackage{algorithm}
\usepackage{algpseudocode}
\usepackage{mysymbol}
\usepackage{enumitem}
\usepackage[algo2e]{algorithm2e}
\usepackage{amsmath}
\allowdisplaybreaks

 \makeatletter
\def\BState{\State\hskip-\ALG@thistlm}
\makeatother

\makeatletter
\newcommand{\algmargin}{\the\ALG@thistlm}
\makeatother
\newlength{\whilewidth}
\algdef{SE}[parWHILE]{parWhile}{EndparWhile}[1]
  {\parbox[t]{\dimexpr\linewidth-\algmargin}{%
     \hangindent\whilewidth\strut\algorithmicwhile\ #1\ \algorithmicdo\strut}}{\algorithmicend\ \algorithmicwhile}%
\algnewcommand{\parState}[1]{\State%
\parbox[t]{\dimexpr\linewidth-\algmargin}{\strut #1\strut}}

\makeatletter
\def\Let@{\def\\{\notag\math@cr}}
\makeatother

\newcommand\numberthis{\addtocounter{equation}{1}\tag{\theequation}}

\usepackage[hidelinks,backref=page]{hyperref}
\definecolor{darkred}{RGB}{150,0,0}
\definecolor{darkgreen}{RGB}{0,150,0}
\definecolor{darkblue}{RGB}{0,0,150}
\hypersetup{colorlinks=true, linkcolor=darkred, citecolor=darkblue, urlcolor=darkblue}

\renewcommand*{\backref}[1]{}
\renewcommand*{\backrefalt}[4]{%
    \ifcase #1 (Not cited.)%
    \or        (Cited on page~#2.)%
    \else      (Cited on pages~#2.)%
    \fi}
    
\title{\textbf{Universal Adversarial Directions}}

\date{}

\author{
Ching~Lam~Choi\thanks{Department of Computer Science and Engineering, The Chinese University of Hong Kong, clchoi1@cse.cuhk.edu.hk} , 
Farzan~Farnia\thanks{Department of Computer Science and Engineering, The Chinese University of Hong Kong, farnia@cse.cuhk.edu.hk}
	}

\begin{document}
\maketitle

\begin{abstract}Despite their great success in image recognition tasks, deep neural networks (DNNs) have been observed to be susceptible to universal adversarial perturbations (UAPs) which perturb all input samples with a single perturbation vector. However, UAPs often struggle in transferring across DNN architectures and lead to challenging optimization problems. In this work, we study the transferability of UAPs by analyzing equilibrium in the \emph{universal adversarial example game} between the classifier and UAP adversary players. We show that under mild assumptions the universal adversarial example game lacks a pure Nash equilibrium, indicating UAPs' suboptimal transferability across DNN classifiers. To address this issue, we propose \emph{Universal Adversarial Directions (UADs)} which only fix a universal direction for adversarial perturbations and allow the perturbations' magnitude to be chosen freely across samples. We prove that the UAD adversarial example game can possess a Nash equilibrium with a pure UAD strategy, implying the potential transferability of UADs. We also connect the UAD optimization problem to the well-known principal component analysis (PCA) and develop an efficient PCA-based algorithm for optimizing UADs. We evaluate UADs over multiple benchmark image datasets. Our numerical results show the superior transferability of UADs over standard gradient-based UAPs. \end{abstract}

\section{Introduction}Deep neural networks (DNNs) have achieved great success in many supervised learning problems from computer vision \cite{krizhevsky2012imagenet}, speech recognition \cite{deng2013new}, natural language processing \cite{otter2020survey}, and computational biology \cite{wainberg2018deep}. Their performance, however, has been observed to be highly susceptible to small perturbations to the neural network's input data widely recognized as \emph{adversarial attacks} \cite{szegedy2013intriguing,biggio2013evasion,goodfellow2014explaining}. A typical adversarial attack scheme assigns a norm-bounded perturbation to an input feature vector, where the designed perturbation is intended to fool either a known DNN (white-box adversarial attacks) or an unknown DNN (black-box adversarial attacks) to predict a wrong label. Over the recent years, adversarial attack and robust training schemes have received enormous attention in the machine learning community.

An adversarial attack scheme typically designs different perturbation vectors for different input data. This property allows the attack algorithm to tailor the designed perturbation to every specific input sample and further empowers the adversary to attain higher success rates in misleading a DNN machine. On the other hand, the influential study by \cite{moosavi2017universal} has empirically shown the existence of a \emph{universal adversarial perturbation (UAP)} that can change the target classifier's predictions over a significant fraction of input samples. As demonstrated in this work and several other papers on universal perturbations \cite{hendrik2017universal,mopuri2017fast,neekhara2019universal,behjati2019universal}, while UAPs are highly constrained across different input data, they still manage to achieve a fair success rate on unseen test data. 

While UAPs can successfully attack a target DNN machine, the recent papers \cite{hashemi2020transferable,park2021distinctive} have reported that UAPs generated by standard gradient-based methods could weakly transfer to unobserved DNN classifiers different from the source DNN used for their construction. Specifically, the reported results suggest that gradient-based UAPs perform noticeably weaker than standard PGD adversarial perturbations in transferring to an unseen DNN architecture. 
Such observations motivate the following question:\vspace{1mm}

{Why do gradient-based UAPs perform suboptimally in transferring to different DNN classifiers?}

The answer to the above question will play a key role in understanding and improving the generalization and transferability properties of universal perturbations. In this work, we focus on the above question and apply a max-min approach to examine the transferability features of UAPs. The applied max-min framework extends the adversarial example game introduced by \cite{bose2020adversarial} for generating transferable adversarial examples to the setting of universal adversarial perturbations. According to the adversarial example game, the adversary attempts to find an attack strategy for generating adversarial examples that achieves the maximum success rate against the most robust classifier from a given function space. We show that under some mild assumptions on a DNN architecture, every universal attack scheme can be completely thwarted by the classifier player. From a game-theoretic perspective, the universal adversarial example game possesses no pure Nash equilibria where the players' deterministic strategies are simultaneously optimal. Consequently, while a gradient-based UAP can significantly drop the performance of a fixed target DNN, the same UAP may have limited impact on a modified DNN function. 

To study and address the transferability suboptimality of gradient-based UAPs, we introduce a variant of universal adversarial perturbations which we call \emph{Universal Adversarial Directions (UADs)}. According to the UAD approach, the adversary is only constrained to generate the perturbations along a unique direction in the sample space. Therefore, the UAD perturbations are no longer required to share the same magnitude and could be chosen differently for different input samples. In particular, unlike gradient-based UAPs, the UAD adversary has the freedom of choosing not to perturb an input sample, which sounds a sensible option in the evaluation of a universal adversarial attack scheme.

In order to find an effective UAD, we introduce a bilevel max-max optimization problem and propose a projected gradient-based algorithm to find a stationary solution in its optimization landscape. Moreover, we develop an efficient principal component analysis (PCA)-based approach to approximate the solution to the UAD optimization problem. The PCA-based method indeed finds the top principle component of the matrix of unnormalized fast gradient method (FGM) perturbations to training data. We provide a stochastic optimization algorithm for computing the solution to the PCA-based optimization problem that is suitable for large-scale machine learning problems. 

We perform theoretical analysis of the equilibrium properties of the UAD adversarial example game. We show that under the assumption that the Fast Gradient Method (FGM)-perturbation matrix has a unique top principal component, the PCA-based approximate UAD game will possess a  Nash equilibrium with a pure strategy for the universal adversary. This result indicates the existence of a single UAD with the maximum impact on the most robust classifier. In the general case, our analysis suggests an extension of the UAD framework to rank-constrained adversarial attacks where the designed perturbations are restricted to a low-rank linear subspace. We show that the rank-constrained adversarial example game will generally possess a Nash equilibrium with a pure adversarial attack strategy. 


Finally, we discuss the results of several numerical experiments comparing the performance of UAPs and UADs over standard image datasets and DNN architectures. Our experimental results support the better transferability and generalizability of UADs  over gradient-base UAPs. In addition, the numerical results suggest that the designed UAD can be applied as a universal perturbation with a similar or even better performance than gradient-based UAP attack schemes. 
We can summarize the main contributions of our work as follows:
\begin{itemize}[leftmargin=7mm]
    \item Analyzing the transferability of universal adversarial perturbations through the max-min framework of adversarial example games
    \item Proposing universal adversarial directions (UADs) as an extension of universal adversarial perturbations
    \item Proving the existence of Nash equilibiria with a pure UAD attack strategy in universal adversarial direction games
    \item Developing an efficient PCA-based algorithm for optimizing UADs
    \item Conducting an empirical study of the performance of UADs compared to gradient-based UAPs.
\end{itemize}

\subsection{Related Work}

Since their introduction by \cite{moosavi2017universal}, universal adversarial attacks have been extensively studied in the machine learning literature. The related literature includes a large body of papers on generating universal perturbations \cite{khrulkov2018art,hayes2018learning,poursaeed2018generative,hirano2020simple,hashemi2020transferable,zhang2021universal}, \revise{black-box universal perturbations \cite{tsuzuku2019structural,tursynbek2021adversarial}}, and on defense methods against universal adversarial attacks \cite{akhtar2018defense,shafahi2018universal,mummadi2019defending}. In our work, we focus on the gradient-based universal adversarial perturbations maximizing the perturbed loss function, which as discussed by \cite{shafahi2018universal} nicely connects to the bilevel optimization problem of universal adversarial example games. We note that the iterative deepfool-based approach in \cite{moosavi2017universal}, the singular vector-based approach in \cite{khrulkov2018art}, and generative model-based method in \cite{hashemi2020transferable} are indeed different from our analyzed gradient-based UAPs which better match our formulation of UAD optimization problems.  

In addition, the equilibrium and convergence properties of adversarial example games have been analyzed in several recent papers. The related works \cite{bose2020adversarial,meunier2021mixed,zhang2021based} focus on the max-min framework of designing transferable adversarial perturbations. Specifically, \cite{meunier2021mixed} prove that the standard adversarial example game generally lacks pure Nash equilibria and proposes finding the mixed Nash equilibria of the adversarial example game. Also, the analysis by \cite{pal2020game} focuses on the adversarial training game with standard adversarial attacks. We note that the mentioned papers focus on the standard adversarial attack setting which does not directly apply to universal perturbations. On the other hand, the related papers \cite{shafahi2018universal,perolat2018playing} focus on the sequential game of universal adversarial training where the classifier moves first followed by the universal adversary. While this sequence leads to robust classifiers against universal perturbations, it does not address the max-min game of transferable universal perturbations which we discuss in our work.  

In another related work, \cite{khrulkov2018art} use the singular vectors of the DNN's Jacobian matrices at different layers as universal  perturbations. On the other hand, our approximate UAD framework chooses the top right-singular vector of the DNN loss's gradient with respect to training data as a \textit{universal adversarial direction} which allows optimizing the perturbations' magnitudes unlike \cite{khrulkov2018art}'s proposed approach. Similarly, the SVD-based approach by \cite{deshpande2019universal} uses the singular vectors of normalized FGSM and PGD perturbations as UAPs and does not focus on the single-direction UAD attacks and its game-theoretic aspects. In addition, we theoretically analyze equilibrium in the UAD adversarial example game. Finally, we note that the SVD-based analyses in \cite{moosavi2017universal,moosavi2017robustness} target the data matrix's singular vectors  which is different from our work's PCA-based analysis of the loss's gradient matrix.
\section{Preliminaries}
In this section, we review some standard definitions and tools regarding standard and universal adversarial attacks. Throughout the paper, we consider a supervised learning setting where the goal is to predict a label variable $Y\in\mathcal{Y}$ from the observation of a $d$-dimensional feature vector $\mathbf{X}\in\mathcal{X}\subseteq \mathbb{R}^d$. Given a loss function $\ell(y,y')$ for labels $y$ and $y'$, the standard empirical risk minimization (ERM) learner aims to find a classifier function $f\in\mathcal{F}$ minimizing the expected prediction loss over a given function space $\mathcal{F}$. 

However, the ERM learner has been observed to lack robustness against norm-bounded adversarial perturbations. To generate a standard norm-bounded adversarial perturbation for classifier $f$, input $(\mathbf{x},y)$, attack norm $\Vert \cdot
\Vert$, and attack power $\epsilon\ge 0$, the adversary finds an $\epsilon$-norm-bounded perturbation  $\boldsymbol{\delta}\in\mathbb{R}^d$ maximizing the prediction loss for input $\mathbf{x},y$:
\begin{equation}\label{Eq: Standard Adversarial Perturbation}
    \max_{\boldsymbol{\delta}:\: \Vert \boldsymbol{\delta}\Vert \le \epsilon}\; \ell\bigl( f(\mathbf{x}+\boldsymbol{\delta}) , y \bigr).
\end{equation}
In our theoretical analysis, we choose the attack norm function as the standard $L_2$ (Euclidean) norm. Note that the perturbation designed by solving \eqref{Eq: Standard Adversarial Perturbation} is a function of input data $(\mathbf{x},y)$, which can result in different perturbations for different input samples. 

On the other hand, a universal adversarial perturbation (UAP) adds the same perturbation to all input data points. Given $n$ training samples $(\mathbf{x}_i,y_i)_{i=1}^n$, a standard approach to design a UAP is through the following optimization problem maximizing the averaged prediction loss for the universally-perturbed training data:
\begin{equation}\label{Eq: Universal Adversarial Perturbation}
    \max_{\boldsymbol{\delta}:\: \Vert \boldsymbol{\delta}\Vert \le \epsilon}\; \frac{1}{n}\sum_{i=1}^n\ell\bigl( f(\mathbf{x}_i+\boldsymbol{\delta}) , y_i \bigr).
\end{equation}
Given an adversarial attack scheme, an adversarial training method trains the classifier using the generated adversarial examples. As a result, the standard adversarial training method \cite{madry2017towards} solves the following min-max optimization problem where the perturbations are generated separately for different input data:
\begin{align}\label{Eq:Standard Min_max AT}
   \min_{f\in\mathcal{F}}\; \frac{1}{n}\sum_{i=1}^n\biggl[\,\max_{\boldsymbol{\delta}_i:\: \Vert \boldsymbol{\delta}_i \Vert\le \epsilon}\, \ell\bigl(\, f(\mathbf{x}_i+\boldsymbol{\delta}_i) , y_i\, \bigr) \,\biggr] \: \equiv \: \min_{f\in\mathcal{F}}\; \max_{\substack{\boldsymbol{\delta}_1,\ldots,\boldsymbol{\delta}_n: \\
   \forall i, \: \Vert \boldsymbol{\delta}_i \Vert\le \epsilon}
   }\; \frac{1}{n}\sum_{i=1}^n\bigl[ \ell\bigl(\, f(\mathbf{x}_i+\boldsymbol{\delta}_i) , y_i\, \bigr) \bigr] \numberthis
\end{align}
To perform universal adversarial training through UAPs, \cite{shafahi2018universal}  introduce the following min-max optimization problem:
\begin{equation}\label{Eq:Universal Min_max AT}
   \min_{f\in\mathcal{F}}\; \max_{\boldsymbol{\delta}:\: \Vert\boldsymbol{\delta} \Vert\le \epsilon}\; \frac{1}{n}\sum_{i=1}^n\bigl[ \ell\bigl(\, f(\mathbf{x}_i+\boldsymbol{\delta}) , y_i\, \bigr) \bigr]. 
\end{equation}
Note that the standard and universal adversarial training problems have different maximization variables, where the maximization variable in the standard adversarial training problem \eqref{Eq:Standard Min_max AT} has a size dependent on the training set size $n$, while the the maximization variable in the universal adversarial training \eqref{Eq:Universal Min_max AT} is independent of the number of training examples.

\section{Universal Adversarial Example Games}
In this section, we aim to analyze the transferability features of UAPs. To do this, we extend the adversarial example game framework introduced by \cite{bose2020adversarial} to the setting of universal perturbations. This extension, which we call the \emph{universal adversarial example game}, is based on the following max-min optimization problem searching for the most transferable norm-bounded UAP $\boldsymbol{\delta}\in\mathbb{R}^d$ against the most robust classifier over function space $\mathcal{F}$:
\begin{equation}\label{Eq: UAP Game}
    \max_{\boldsymbol{\delta} :\: \Vert\boldsymbol{\delta}\Vert \le \epsilon}\; \min_{f\in\mathcal{F}} \; \frac{1}{n}\sum_{i=1}^n\ell\bigl( f(\mathbf{x}_i+\boldsymbol{\delta}) , y_i \bigr).
\end{equation}
Note that the above bilevel optimization problem represents a zero-sum game where the UAP player designing an $\epsilon$-norm-bounded universal perturbation $\boldsymbol{\delta}\in\mathbb{R}^d$ moves first followed by the classifier player $f\in\mathcal{F}$ predicting the label from the universally-perturbed feature vector. The solution to this max-min problem provides the most transferable universal perturbation with the highest worst-case impact on the classifiers in $\mathcal{F}$. 

Also, we highlight the difference between the above max-min optimization problem and the min-max problem of universal adversarial training \cite{shafahi2018universal,perolat2018playing}. Although these two problems only differ in the order of minimization and maximization, they do not necessarily share the same solution as the game could lack a pure Nash equilibrium where the deterministic minimization and maximization strategies are simultaneously optimal. In the following theorem, we indeed show that under a mild assumption on classifier space $\mathcal{F}$ which applies to multi-layer DNN architectures, every universal perturbation achieves the same transferability score against the robust classifier, revealing that the min-max and max-min problems have different solutions. 
\begin{theorem}\label{Thm: UAP Nash}
Suppose that for every $f\in\mathcal{F}$ and bias vector $\mathbf{b}\in\mathbb{R}^d$ the function $f_{\mathbf{b}}:\mathbb{R}^d\rightarrow \mathbb{R}$ defined as $f_{\mathbf{b}}(\mathbf{x}):= f(\mathbf{x} + \mathbf{b}) $ still belongs to $\mathcal{F}$. Then, 
\begin{itemize}[leftmargin=5mm]
    \item The minimized objective function in \eqref{Eq: UAP Game} over function space $\mathcal{F}$ takes the same value for every choice of $\boldsymbol{\delta}\in\mathbb{R}^d$.
    \item The universal adversarial example game has no Nash equilibria with a non-zero pure strategy $\boldsymbol{\delta}^*\neq \mathbf{0}$ for the UAP adversary.
\end{itemize}
\end{theorem}
\begin{proof}
    We defer the proof to the Appendix.
\end{proof}
Theorem \ref{Thm: UAP Nash} observes that if the classifier set $\mathcal{F}$ is closed under the addition of an input bias vector, then the effect of a UAP can be reversed by subtracting the UAP using the bias vector. Hence, one can expect that a UAP could struggle in transferring to other DNN classifiers, since its effect is reversible. In next section, we discuss how to improve the performance of universal perturbations by addressing the reversibility of UAPs in the adversarial example game.

\section{Universal Adversarial Directions}

To address the lack of pure Nash equilibria in the universal adversarial example game, we propose a modified notion of universal perturbations. One of the main factors preventing the UAP game from reaching an equilibrium is the UAP adversary's restriction to apply the same perturbation to all input data. As a result,  a classifier that knows the UAP in advance can easily reverse the effect of the fixed perturbation. Based on this discussion, we propose considering a universal adversary capable of choosing between adding the universal perturbation or not adding the perturbation for every individual sample. Such a universal adversarial attack is therefore only constrained to generate all the perturbations along the same direction. The discussion motivates the definition of a \emph{universal adversarial direction}.

\begin{mydef}
We call a unit-norm $\boldsymbol{\delta}$ a universal adversarial direction (UAD) if the adversarial perturbation $\delta(\mathbf{x},y)$ designed for every input $(\mathbf{x},y)$ is aligned with $\boldsymbol{\delta}$, i.e. $\delta(\mathbf{x},y) = \tau_{\mathbf{x},y}\boldsymbol{\delta}$ holds for a scalar $\tau_{\mathbf{x},y}\in\mathbb{R}$. 
\end{mydef}

To generate a powerful UAD with the maximum impact on a given classifier $f$, we propose solving the following optimization problem: 
\begin{align}\label{Eq: UAD max_max constrained optimization}
    \max_{\boldsymbol{\delta}:\: \Vert \boldsymbol{\delta}\Vert \le 1}\; \frac{1}{n}\sum_{i=1}^n \biggl[ \max_{\tau_i\in\mathbb{R}:\, \vert \tau_i\vert\le \epsilon }\: \ell\bigl(f(\mathbf{x}_i + \tau_i\boldsymbol{\delta}),y_i \bigr)\biggr] \: \equiv \: \max_{\substack{\boldsymbol{\delta},\tau_1,\ldots,\tau_n: \\
     \Vert \boldsymbol{\delta}\Vert \le 1,\, \forall i:\, |\tau_i|\le \epsilon}}\; \frac{1}{n}\sum_{i=1}^n \bigl[ \ell\bigl(f(\mathbf{x}_i + \tau_i\boldsymbol{\delta}),y_i \bigr)\bigr]\numberthis 
\end{align}
In the above formulation, every scalar variable $\tau_i$  represents the magnitude of the additive perturbation $\tau_i \boldsymbol{\delta}$ for the $i$th data point $(\mathbf{x}_i,y_i)$. Note that all the perturbations are constrained to be along the optimization variable $\boldsymbol{\delta}$. Taking a standard gradient-based approach to optimize the UAD-based objective function in \eqref{Eq: UAD max_max constrained  optimization}, one can apply the projected gradient method (PGM). Here, the optimization variables $\boldsymbol{\delta},\tau_1,\ldots,\tau_n$ are optimized using the projected gradient ascent algorithm. To derive a stochastic version of the optimization algorithm using a mini-batch of training data at every iteration, we propose Algorithm \ref{alg: UADgrad} applying stochastic projected gradient ascent to solve the UAD  problem.

\begin{figure}
\begin{algorithm}[H]
\textbf{Initialize} perturbation $\boldsymbol{\delta}^0$, stepsizes $\eta_1, \eta_2$, batch-size $B$, number of inner updates $K$\\
\For{$t = 0, \cdots, T-1$}{
    \textbf{Draw} a mini-batch of samples $(\mathbf{x}_{t_i},y_{t_i})_{i=1}^B$ \vspace{.15cm}\\
    \textbf{Initialize}  magnitudes $\tau^0_1,\ldots,\tau^0_B$\vspace{.15cm} \\ 
    \For{$k = 0, \cdots, K-1$}{
    \vspace{-.2cm}
    \begin{flalign} 
        &\forall i:\; \tau^{k+1}_{i} = \tau^{k}_{i} + \eta_2 \frac{\text{\rm d} \ell\bigl(f(\mathbf{x}_{t_i} + \tau^{k}_{i} \boldsymbol{\delta}^t),y_{t_i}\bigr)}{\text{\rm d} \tau}& \\
        &\forall i:\; \tau^{k+1}_{i} = \min\{\max\{\tau^{k+1}_{i},-\epsilon\},\epsilon \}&\nonumber
    \end{flalign}
    }\vspace{-.7cm}
     \begin{flalign} 
       &\boldsymbol{\delta}^{t+1} = \boldsymbol{\delta}^{t} + \frac{\eta_1}{B}\sum_{i=1}^B \nabla_{\boldsymbol{\delta}}  \ell\bigl(f(\mathbf{x}_{t_i} + \tau^{K}_{i} \boldsymbol{\delta}^t),y_{t_i}\bigr)& \\
     &\boldsymbol{\delta}^{t+1} = \frac{\boldsymbol{\delta}^{t+1}}{\max\{1,\Vert \boldsymbol{\delta}^{t+1}\Vert\}}& \nonumber
    \end{flalign}\vspace{-.2cm}
}
\textbf{Output} $\boldsymbol{\delta}=\boldsymbol{\delta}^T$
\caption{UAD-Projected Gradient Ascent}\label{alg: UADgrad} 
\end{algorithm}
\end{figure}
\vspace{-3mm}


\vspace{.14cm}
\section{A PCA-based Approach to Universal Adversarial Directions}

In the previous section, we defined UADs and introduced a gradient-based algorithm for solving the UAD optimization problem. However, since the underlying UAD optimization task maximizes a non-concave objective function, the algorithm is only guaranteed to find a first-order stationary solution under regularity assumptions. In this section, we use a Taylor series-based approximation of the UAD optimization objective to relate the optimal UAD to the top principal component of the fast gradient method (FGM) perturbation matrix. {This connection results in an analytically tractable optimization problem for approximating the optimal UAD, which facilitates the analysis of UADs.}

To build the connection, we focus on the following Lagrangian version of the UAD optimization problem for a coefficient $\lambda>0$ replacing the role of attack power $\epsilon$ in the UAD problem:
\begin{equation}\label{Eq: UAD Lagrangian optimization}
    \max_{\boldsymbol{\delta}:\: \Vert \boldsymbol{\delta}\Vert \le 1}\; \frac{1}{n}\sum_{i=1}^n \biggl[ \max_{\tau_i\in\mathbb{R} }\: \ell\bigl(f(\mathbf{x}_i + \tau_i\boldsymbol{\delta}),y_i \bigr) - \frac{\lambda}{2} \tau_i^2\biggr].
\end{equation}
\vspace{-3mm}
\begin{prop}\label{Prop: PCA approximation}
Suppose that $\ell \circ f$ is a $\rho$-smooth differentiable function of the input feature vector $\mathbf{x}$, i.e. for every $\mathbf{x},\mathbf{x}',y$ we have $\Vert \nabla_\mathbf{x}\ell(f(\mathbf{x}),y)- \nabla_\mathbf{x}\ell(f(\mathbf{x}'),y)\Vert\le \rho \Vert \mathbf{x}-\mathbf{x}'\Vert$. Assuming that $\Vert \boldsymbol{\delta}\Vert^2\le B$ holds with probability $1$ and $\lambda > B\rho$, the following inequalities hold for every sample $(\mathbf{x}_i,y_i)$:
\begin{align}
    \frac{1}{2(\lambda + B\rho)} \bigl(\boldsymbol{\delta}^\top \nabla_{\mathbf{x}}\ell(f(\mathbf{x}_i ),y_i)\bigr)^2 \; &\le \; \max_{\tau_i\in\mathbb{R} }\left\{ \ell\bigl(f(\mathbf{x}_i + \tau_i\boldsymbol{\delta}),y_i \bigr) - \frac{\lambda}{2} \tau_i^2 \right\} - \ell(f(\mathbf{x}_i),y_i )  \\    
     \; &\le\;  \frac{1}{2(\lambda - B\rho)} \bigl(\boldsymbol{\delta}^\top \nabla_{\mathbf{x}}\ell(f(\mathbf{x}_i ),y_i)\bigr)^2 
\end{align}
\end{prop}
\begin{proof}
We defer the proof to the Appendix.
\end{proof}
The above proposition suggests optimizing the above upper-bound on the UAD optimization objective function approximating the objective function within an error factor of $\frac{\lambda + \rho}{\lambda - \rho}$:  
\begin{align}\label{Eq: UAD Lagrangian optimization PCA approx}
    &\max_{\boldsymbol{\delta}:\: \Vert \boldsymbol{\delta}\Vert \le 1}\; \frac{1}{n}\sum_{i=1}^n \biggl[ \ell(f(\mathbf{x}_i),y_i ) + \frac{\bigl(\boldsymbol{\delta}^\top \nabla_{\mathbf{x}}\ell(f(\mathbf{x}_i ),y_i)\bigr)^2}{2(\lambda - \rho)} \biggr]\nonumber \\
   \equiv \;&\frac{1}{n}\sum_{i=1}^n \bigl[ \ell(f(\mathbf{x}_i),y_i )\bigr] + \frac{1}{2(\lambda - \rho)}\max_{\boldsymbol{\delta}:\: \Vert \boldsymbol{\delta}\Vert \le 1}\biggl\{ \boldsymbol{\delta}^\top \biggl(\frac{1}{n}\sum_{i=1}^n\nabla_{\mathbf{x}}\ell(f(\mathbf{x}_i ),y_i)\nabla_{\mathbf{x}}\ell(f(\mathbf{x}_i ),y_i)^\top\biggr)\boldsymbol{\delta}\biggr\}. 
\end{align}
We observe that the solution to the above optimization problem is indeed the top principal component, i.e. the top right-singular vector, of the following matrix $G_S(f)$ including the loss's gradient for classifier $f$ with respect to training samples in dataset $S=\{(\mathbf{x}_i,y_i)_{i=1}^n\}$:
\begin{equation}
    G_S(f) := \frac{1}{\sqrt{n}}\begin{bmatrix} \, \nabla_{\mathbf{x}}\ell\bigl(f(\mathbf{x}_1 ),y_1\bigr)\,\vspace{1mm} \\
    \vdots\vspace{1mm} \\
   \, \nabla_{\mathbf{x}}\ell\bigl(f(\mathbf{x}_n ),y_n\bigr)\,
\end{bmatrix}_{n\times d}
\end{equation}
The above matrix contains the unnormalized fast gradient method (FGM) perturbations as its rows. Note that if we perform a similar first-order approximation analysis for the UAP optimization, the approximate solution will be the mean of the rows of the above matrix. Therefore, according to the above first-order analysis, the UAD framework approximately substitutes the average row of the loss's gradients used by the UAP approach with the gradient matrix's top principal component, which could better capture the existing structures in the FGM-perturbation matrix $G_S(f)$. {We note that the tractability of the PCA-based approach is due to the choice of $\ell_2$-norm for the universal direction. For other $\ell_p$-norm functions, the computation of the optimal universal direction could be intractable.}

Inspired by several recent works applying stochastic optimization methods for computing the top singular vector \cite{shamir2015stochastic,shamir2016convergence}, we propose Algorithm \ref{alg: UADpca} to compute the PCA-based approximation of the optimal UAD. In particular, the stochastic nature of Algorithm \ref{alg: UADpca} suits large-scale machine learning problems where a direct application of the singular value decomposition (SVD) algorithm could be computationally difficult.     


\begin{figure}
\begin{algorithm}[H]
\textbf{Initialize} perturbation $\boldsymbol{\delta}^0$, stepsize $\eta$, batch-size $B$\\
\For{$t = 0, \cdots, T-1$}{
    \textbf{Draw} a mini-batch of samples $(\mathbf{x}_{t_i},y_{t_i})_{i=1}^B$\\
     \vspace{-0.3cm}\begin{flalign} 
       &\boldsymbol{\delta}^{t+1} = \boldsymbol{\delta}^{t} + \frac{\eta}{B}\sum_{i=1}^B \biggl[\left({\boldsymbol{\delta}^t}^\top\nabla_{\mathbf{x}}  \ell\bigl(f(\mathbf{x}_{t_i}),y_{t_i}\bigr)\right)\nabla_{\mathbf{x}}  \ell\bigl(f(\mathbf{x}_{t_i}),y_{t_i}\bigr)\biggr]& \nonumber \\
     &\boldsymbol{\delta}^{t+1} = \frac{\boldsymbol{\delta}^{t+1}}{\max\{1,\Vert \boldsymbol{\delta}^{t+1}\Vert\}}& \nonumber
    \end{flalign}\vspace{-.2cm}
}
\textbf{Output} $\boldsymbol{\delta}=\boldsymbol{\delta}^T$
\caption{UAD-Principal Component Analysis}\label{alg: UADpca}
\end{algorithm}
\end{figure}

\captionsetup[figure]{font=normalsize}
\captionsetup[table]{font=normalsize}


\section{Nash Equilibria in UAD Games}

We previously discussed that UAPs suffer from the lack of equilibria in the universal adversarial example game. In this section, our aim is to show that a similar zero-sum game adapted for our proposed UADs will indeed possess a non-trivial Nash equilibrium, where a fixed non-zero direction is the most effective UAD against any classifier in function space $\mathcal{F}$ the most. To prove such a guarantee, we first define the \emph{universal adversarial direction game} played between an adversary player searching for the most effective direction $\boldsymbol{\delta}\in\mathbb{R}^d$, along which the designed perturbations can mislead the classifier player $f\in\mathcal{F}$. Mathematically, we use the following max-min optimization problem for universal adversarial direction games:
\begin{equation}\label{Eq: UAD Game}
    \max_{\Vert\boldsymbol{\delta}\Vert \le 1}\; \min_{f\in\mathcal{F}} \; \frac{1}{n}\sum_{i=1}^n\biggl[\max_{\tau_i\in\mathbb{R}: \, \vert \tau_i\vert\le\epsilon } \ell\bigl(f(\mathbf{x}_i + \tau_i\boldsymbol{\delta}),y_i \bigr) \biggr].
\end{equation}
Following the PCA-based approximation of the UAD optimization problem and defining $L_S(f) = \frac{1}{n}\sum_{i=1}^n \ell\bigl(f(\mathbf{x}_i),y_i \bigr)$ as the averaged prediction loss over unperturbed training data, we can apply Proposition \ref{Prop: PCA approximation} to formulate the approximate universal adversarial direction game with the following optimization problem with parameter $\eta>0$:
\begin{align}\label{Eq: UAD Game Approx}
\max_{\Vert\boldsymbol{\delta}\Vert \le 1}\, \min_{f\in\mathcal{F}} \, \frac{1}{n}\sum_{i=1}^n\biggl[ \ell\bigl(f(\mathbf{x}_i),y_i \bigr) + \eta \bigl(\boldsymbol{\delta}^\top\nabla_{\mathbf{x}}\ell\bigl(f(\mathbf{x}_i ),y_i\bigr)\bigr)^2\biggr] \; =\; \max_{\Vert\boldsymbol{\delta}\Vert \le 1}\, \min_{f\in\mathcal{F}} \, L_S(f) + \eta \boldsymbol{\delta}^\top G_S(f)G_S(f)^\top\boldsymbol{\delta}.
\end{align}

Therefore, the min-max problem corresponding to the above approximate UAD optimization task reduces to the following one-level optimization problem where $\Vert\cdot\Vert_{2}$ denotes the $L_2$ operator norm:
\begin{align}\label{Eq: UAD Min-Max Game Approx}
\min_{f\in\mathcal{F}} \, \max_{\Vert\boldsymbol{\delta}\Vert \le 1}\,  L_S(f) + \eta \boldsymbol{\delta}^\top G_S(f)G_S(f)^\top\boldsymbol{\delta} \; \equiv\;    \min_{f\in\mathcal{F}} \,  L_S(f)+ \eta \bigl\Vert G_S(f)\bigr\Vert^2_{2}.
\end{align}
The following theorem proves that if for every minimizer $f^*\in\mathcal{F}$ of the above objective function, the matrix $G_S(f^*)$ has a unique top singular value, then the approximate universal adversarial example game will possess a Nash equilibrium with a pure strategy for the UAD adversary.
\begin{theorem}\label{Thm: UAD Nash}
Define the approximate UAD objective function $\mathcal{V}(f) := L_S(f)+ \eta \bigl\Vert G_S(f)\bigr\Vert^2_{2}$. Suppose that the matrix set $\{L_S(f)I_d + \eta G_S(f)G_S(f)^\top: f\in\mathcal{F}\}$, where $I_d$ denotes the identity matrix, is convex and compact. Then,
\begin{itemize}[leftmargin=7mm]
    \item If for every minimizer $f^*\in\mathcal{F}$ of $\mathcal{V}(f)$ the matrix $G_S(f^*)$ has a unique top right-singular vector, there exists a Nash equilibrium to the approximate universal adversarial example game with a pure strategy $\boldsymbol{\delta^*}\in\mathbb{R}^d$ for the UAD adversary.
    \item If for every minimizer $f^*\in\mathcal{F}$ of $\mathcal{V}(f)$ the matrix $G_S(f^*)$ has the top singular value with multiplicity at most $r$, the approximate universal adversarial example game has a mixed Nash equilibrium where the UAD player always chooses the adversarial direction from a universal $r$-dimensional space $\Delta_r\in\mathbb{R}^{r\times d}$ spanned by a group of $r$ universal vectors  $\{\boldsymbol{\delta}^*_1,\ldots,\boldsymbol{\delta}^*_r \}$. 
\end{itemize}
\end{theorem}
\begin{proof}
We defer the proof to the Appendix.
\end{proof}

The above theorem shows that unlike the UAP adversarial example game, the UAD-based game can indeed possess Nash equilibria with a pure or in general rank-constrained strategy for the universal adversary player. Hence, Theorem \ref{Thm: UAD Nash} indicates that the UAD adversary can apply a non-trivial pure strategy with the maximum impact on the classifier.   

\section{Numerical Results}

\begin{table}[t]
\renewcommand{\arraystretch}{1.10}
    \centering
     \scriptsize
    \begin{tabular}{?c|c?c|c?c|c?c|c?c|c?c|c?c|c?}
    \cline{1-14}
    \hline
    \multicolumn{14}{?c?}{\textbf{Natural and Test Adversarial Accuracies (TA) \& Fooling Rates (FR)}}  \\
    \hhline{|==============|}
    \multicolumn{2}{?c?}{\multirow{2}{*}{\backslashbox{\textbf{Dataset}}{\textbf{Model}}}} & \multicolumn{2}{c?}{ResNet-18} & \multicolumn{2}{c?}{ResNet-34 (50)} & \multicolumn{2}{c?}{DenseNet-121} & \multicolumn{2}{c?}{CaffeNet} & \multicolumn{2}{c?}{VGG-19} & \multicolumn{2}{c?}{\revise{EfficientNet-V2-S}}\\
    \cline{3-14}
    \multicolumn{2}{?c?}{} & TA $\downarrow$ & FR $\uparrow$ & TA $\downarrow$ & FR $\uparrow$ & TA $\downarrow$ & FR $\uparrow$ & TA $\downarrow$ & FR $\uparrow$ & TA $\downarrow$ & FR $\uparrow$ & TA $\downarrow$ & FR $\uparrow$ \\
    \hline
    \multirow{7}{*}{CIFAR-10} &Natural &92.4 & 0.0 &91.9 & 0.0 &92.2 & 0.0 & 85.7 & 0.0 & 87.9 &0.0 & 86.9 & 0.0\\
                              &UAD &\textbf{11.3} & \textbf{88.4} &\textbf{12.6} &\textbf{85.4} &\textbf{10.3} & 86.9 &\textbf{24.7} & \textbf{74.0} &\textbf{27.5} &71.2 &\textbf{27.4} & \textbf{69.4}\\
                              &UAD-mag1 & 15.5 &84.7 & 15.5 &83.9 &12.1 &87.8 &25.9 &73.0 &27.6 &\textbf{71.6} &35.5 & 62.2\\
                              &UAP & 44.7 &59.5 &49.0 &67.2 &16.7 &81.2 &66.0 &31.2 &65.9 &31.9 &40.4 & 56.6\\
                              &GAP & 38.5 &61.1 &39.1 & 59.9 &39.2 &60.2 &61.8 &35.2 &40.2 &57.6 &51.7 & 46.4\\
                              &CDA & 42.8 & 56.3 & 46.0 & 51.8 & 25.6 & 73.8 & 54.1 & 40.9 & 60.7 & 35.0 & 51.8 & 43.5\\
                              & GUAP & 13.5 & 86.5 & 15.2 & 84.1 & 10.8 & \textbf{88.0} & 27.7 & 71.3 & 30.8 & 68.0 & 42.5 & 59.4\\
    \hhline{|==============|}
    \multirow{7}{*}{CIFAR-100} &Natural & 69.2 & 0.0 &70.0 & 0.0 &72.0 & 0.0 &60.1 &0.0 &63.8 &0.0 &57.3 & 0.0\\
                              &UAD &\textbf{4.0} &\textbf{91.9} &8.2 &89.3 & \textbf{4.6} &\textbf{97.5} &\textbf{4.0} &\textbf{95.3} &\textbf{5.9} &\textbf{87.0} &8.9 &84.8\\
                              &UAD-mag1 &7.7 &91.9 &9.6 &89.4 &10.8 &88.2 &4.6 &95.1 &7.4 &91.8 &6.3 &84.6\\
                              &UAP &15.9 &82.7 &29.8 & 67.9 &14.6 & 81.3 &13.2 &80.7 &31.0 &64.8 &19.8 & 70.0\\
                              &GAP &15.9 &83.0 &19.8 &78.1 &19.8 &78.6 &25.7 &70.3  &12.5 &86.5 &19.4 & 77.7\\
                              &CDA &24.5 &72.6 &28.4 &68.1 &16.3 &82.0 &32.3 &58.1 &38.4 &56.3 &19.8 &77.0\\
                              &GUAP &4.8 &90.9 &\textbf{8.0} &\textbf{91.3} &6.1 &93.8 &5.2 &94.3 &6.6 &86.3 &\textbf{7.2} &\textbf{86.9}\\
    \hhline{|==============|}
    \multirow{7}{*}{\shortstack{Tiny- \\ ImageNet}} &Natural & 51.5 & 0.0 & 51.5 & 0.0 & 54.4 &0.0 & 37.4 &0.0 &29.5 &0.0 &36.6 &0.0\\
                              &UAD &\textbf{1.2} & \textbf{98.2} &1.0 &\textbf{98.6} &\textbf{1.6} &\textbf{98.3} &\textbf{3.3} &\textbf{98.1} &\textbf{0.4} &\textbf{99.4} &\textbf{1.3} & 97.4\\
                              &UAD-mag1 &1.5 & 98.1 &2.4 & 97.5 &2.4 &97.8 &4.4 &97.2 &0.8 &98.9 &2.7 &97.8 \\
                              &UAP &3.2 &93.9 &5.2 &95.3 &6.3 &95.1 &4.0 &89.8 &3.1 &96.9 &2.3 &\textbf{98.3}\\
                              &GAP &6.2 & 92.8 &3.4 &96.3 &9.2 & 90.0 &8.4 &87.9 &4.8 & 93.1 &5.1 & 93.2\\
                              & CDA & 4.8 & 94.5 & 3.9 & 95.3 & 7.7 & 91.2 & 9.4 & 85.8 & 4.3 & 93.9 & 3.3 & 97.0\\
                              & GUAP & 1.6 & 97.9 & \textbf{0.9} & 98.0 & 1.8 & 98.3 & 4.1 & 96.5 & 0.8 & 99.3 & 2.6 & 97.2\\
    \hhline{|==============|}
    \multirow{7}{*}{\revise{ImageNet}} &\revise{Natural} &69.8 &0.0 &76.1 &0.0 &74.4 &0.0 &56.5 & 0.0 &74.2 &0.0 & 84.2 & 0.0 \\
                              &UAD &\textbf{4.0} & \textbf{96.7} &\textbf{5.7} &\textbf{94.8} &\textbf{8.9} &\textbf{90.0} &\textbf{4.0} &\textbf{96.3} &4.1 &94.6 & 4.3 & \textbf{95.2}\\
                              &UAD-mag1 &4.9 &93.6 &6.6 &92.2 &11.7 &87.2 &6.1 &92.6 &7.2 &92.2 &6.1 &92.7\\
                              &UAP &19.4 &71.4 &34.6 &56.3 &22.5 &67.9 &18.6 &76.9  &22.5 &79.8 &10.9 & 78.8\\
                              &GAP &14.4 &73.5 &19.9 &69.0 &20.7 &69.3 &12.0 &81.5 &19.2 &84.0 &17.7 &71.4\\
                              & CDA &20.4 &75.2 &35.2 &61.8 &29.3 &68.2 &14.9 &80.1 &22.4 &77.4 &27.1 &67.3\\
                              & GUAP &4.5 &95.2 &5.7 &93.1 &11.5 &86.5 &4.1 &96.2 &\textbf{4.1} &\textbf{96.2} &\textbf{4.3} &94.4\\
    \hline
    \end{tabular}
    \vspace{2mm}
    \caption{UAD, UAD-mag1, UAP, GAP, CDA \& GUAP perturbations' effectiveness and fooling rates. The numbers show adversarial test accuracy (the lower the more effective) and fooling rate (the higher the better).}
    \label{tab:strength}
\end{table}


We performed several numerical experiments to evaluate the UAD perturbations' generalizability and transferability on benchmark image datasets including ImageNet \& TinyImageNet \cite{imagenet_cvpr09}, CIFAR-100 \& CIFAR-10 \cite{krizhevsky2009learning}, and MNIST \cite{lecun1998mnist}. Note that TinyImageNet is a reduced version of standard ImageNet dataset containing 100,000 images from 200 ImageNet classes, with 500 colored images for each class; CIFAR-100 and CIFAR-10 consist of 60,000 colored images from 100 and 10 classes, respectively; MNIST contains 70,000 greyscale handwritten digit images from 10 classes. 

In our experiments, we target multiple widely-used DNN architectures including ResNet-18, ResNet-34 (ResNet-50 for ImageNet) \cite{he2016deep}, DenseNet-121 \cite{huang2017densely}, AlexNet/CaffeNet \cite{krizhevsky2012imagenet}, VGG-19 \cite{simonyan2014very} \revise{and the recent EfficientNet-V2-S \cite{tan2021efficientnetv2}}. The neural network classifiers were trained for 100 epochs using the minibatch gradient descent optimization method with a batch-size of 32, learning rate of 3e-4, and weight decay of 1e-4, which were chosen using cross-validation as detailed in the Appendix. 

We generated universal perturbations using the stochastic optimization methods for UAD, UAD-grad, gradient-based UAP, generative adversarial perturbation (GAP) \cite{poursaeed2018generative}, generalized universal adversarial perturbation (GUAP) \cite{zhang2020generalizing} and cross-domain attack (CDA) \cite{naseer2019cross}, with bounded norms controlled by parameter $\epsilon=0.1 \cdot \mathbb{E}_{\hat{P}}[\Vert \textbf{X} \Vert_2]$ (fraction of the mean $L_2$-norm of clean training samples).

\textbf{Universal Perturbations' Effectiveness.} 
We performed attacks on the aforementioned DNN models with perturbations generated via UAD (PCA-based Algorithm \ref{alg: UADpca}), UAD-grad (gradient-based Algorithm \ref{alg: UADgrad}), UAD-mag1, gradient-based UAP, GAP, CDA and GUAP. UAD-mag1 represents the performance of UADs when directly applied with a constant unit $\tau_i=1$ magnitude onto test samples (therefore not optimized during inference time), which are again more effective than UAPs. In Table \ref{tab:strength}, we used an attack power of $\epsilon=0.1 \cdot \mathbb{E}_{\hat{P}}[\Vert \textbf{X} \Vert_2]$ to compare the strength of different adversarial attacks; we report the resultant test accuracies and fooling rates (\% of labels flipped by the attack) across datasets. 

\begin{figure}[t]
\centering
  \includegraphics[width=\textwidth]{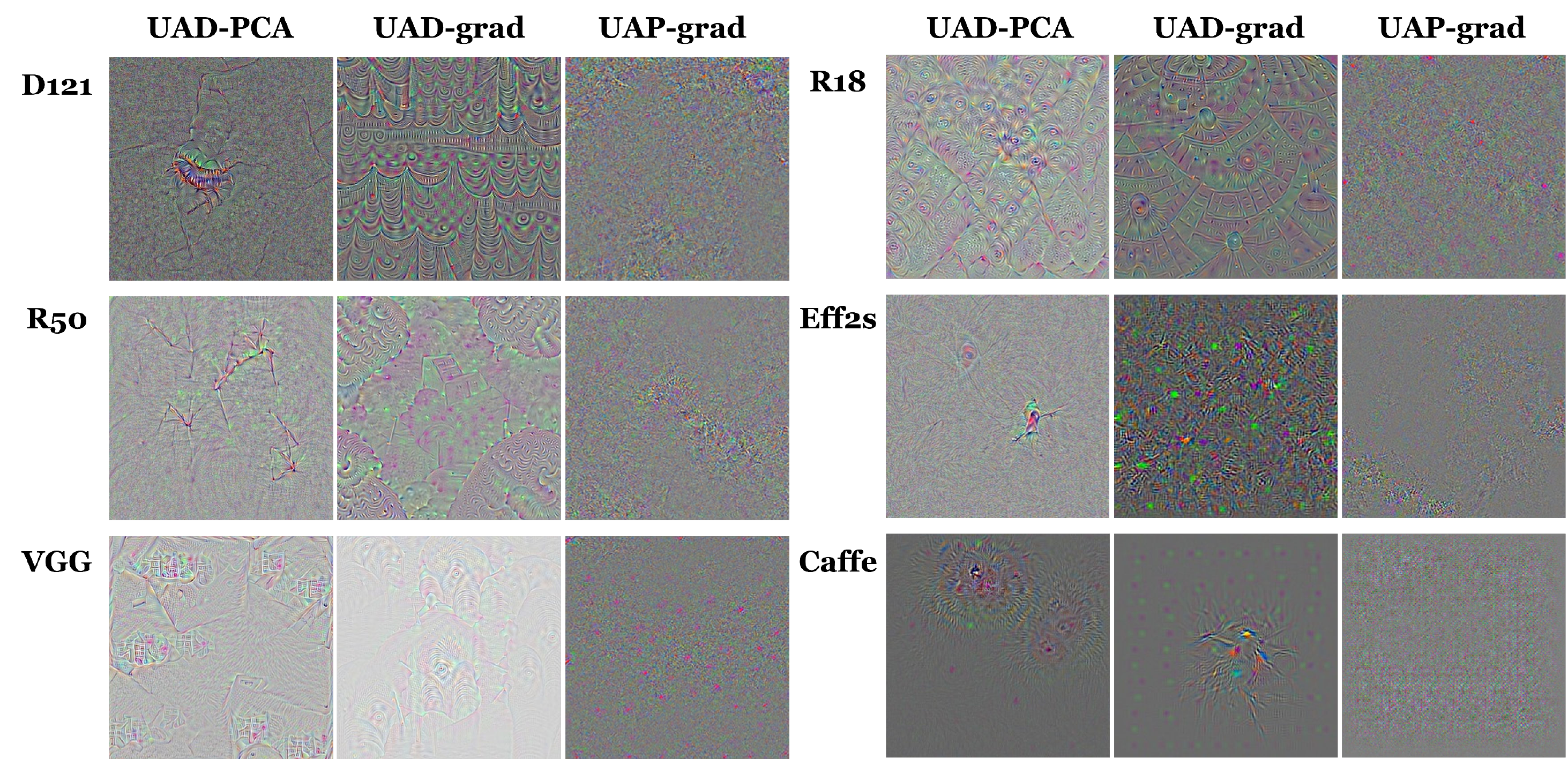}
  \caption{\revise{ImageNet visualizations of UAD-PCA, UAD-grad, UAP noise at $\epsilon=0.1 \cdot \mathbb{E}[\Vert \textbf{X} \Vert_2]$}}
  \label{fig:visuals}
\end{figure}

We further report the fooling rates for UAD-PCA, UAD-grad and UAP perturbations with $\epsilon=.1,.05, .02, .01 \cdot \mathbb{E}_{\hat{P}}[\Vert \textbf{X} \Vert_2]$ in the Appendix Tables \ref{tab:fool_UAD}, \ref{tab:fool_UADG} and \ref{tab:fool_UAP}, for a more fine-grained comparison. In addition to the quantitative scores, we also visualized perturbations generated by the UAD (UAD-PCA), UAD-grad and gradient-based UAP adversaries in Figure \ref{fig:visuals}, with $\epsilon=0.1 \cdot \mathbb{E}[\Vert \textbf{X} \Vert_2]$ adversarial noise on backbone models; in Figure \ref{fig: visuals2}, where horizontal rows correspond to perturbations with decreasing powers ($0.1, 0.05, 0.02, 0.01$). We see that the UAD adversary creates more regular noise patterns with enhanced semantic locality than UAP.

\begin{table}[h!]
\renewcommand{\arraystretch}{1.15}
\begin{subtable}[t]{0.49\linewidth}
    \centering
    \footnotesize
    \begin{tabular}{|l|l|l|l|l|l|l|}
    \hline
    \multicolumn{7}{|c|}{\revise{\textbf{UAD Cosine Similarities for ImageNet}}}  \\
    \hhline{|=======|}
    {\footnotesize \revise{Tar / Src}} & \revise{R18} & \revise{R50} & \revise{D121} & \revise{Alex} & \revise{VGG} & \revise{Eff2s}\\
    \hline
    \revise{R18} & \revise{\textbf{1.00}} & \revise{-0.29} & \revise{0.18} & \revise{-0.29} & \revise{0.21} & \revise{0.25} \\
    \hline
    \revise{R50} & \revise{-0.29} & \revise{\textbf{1.00}} & \revise{0.19} & \revise{-0.24} & \revise{0.35} & \revise{0.30}  \\
    \hline
    \revise{D121} & \revise{0.18} & \revise{0.19} & \revise{\textbf{1.00}} & \revise{0.17} & \revise{-0.14} & \revise{0.13} \\
    \hline
    \revise{Alex} & \revise{-0.29} & \revise{-0.24} & \revise{0.17} & \revise{\textbf{1.00}} & \revise{-0.25} & \revise{0.32} \\
    \hline
    \revise{VGG} & \revise{0.21} & \revise{0.35} & \revise{-0.14} & \revise{-0.25} & \revise{\textbf{1.00}} &\revise{0.31} \\
    \hline
    \revise{Eff2s} & \revise{0.25} & \revise{0.30} &\revise{0.13} &\revise{0.32} &\revise{0.31} &\revise{\textbf{1.00}}\\
    \hhline{|=======|}
    \hline
    \multicolumn{7}{|c|}{\revise{\textbf{UAP Cosine Similarities for ImageNet}}}  \\
    \hhline{|=======|}
    {\footnotesize \revise{Tar / Src}} & \revise{R18} & \revise{R50} & \revise{D121} & \revise{Alex} & \revise{VGG} & \revise{Eff2s}\\
    \hline
    \revise{R18} & \revise{\textbf{1.00}} & \revise{0.00} & \revise{-0.01} & \revise{-0.01} & \revise{-0.01} & \revise{0.00} \\
    \hline
    \revise{R50} & \revise{0.00} & \revise{\textbf{1.00}} & \revise{0.01} & \revise{0.00} & \revise{-0.00} & \revise{0.00}  \\
    \hline
    \revise{D121} & \revise{-0.01} & \revise{0.01} & \revise{\textbf{1.00}} & \revise{-0.01} & \revise{0.00} & \revise{0.00} \\
    \hline
    \revise{Alex} & \revise{-0.01} & \revise{0.00} & \revise{-0.01} & \revise{\textbf{1.00}} & \revise{0.00} & \revise{-0.01} \\
    \hline
    \revise{VGG} & \revise{-0.01} & \revise{-0.00} & \revise{0.00} & \revise{0.00} & \revise{\textbf{1.00}} &\revise{0.00} \\
    \hline
    \revise{Eff2s} & \revise{0.00} & \revise{0.00} &\revise{0.00} &\revise{-0.01} &\revise{0.00} &\revise{\textbf{1.00}}\\
    \hhline{|=======|}
    \end{tabular}
    \caption{Cosine similarity scores for UAD \& UAP on ImageNet.}
    \label{tab:cosine_im}
    \end{subtable}\hfill
    \begin{subtable}[t]{.49\linewidth}
        \centering
        \footnotesize
    \begin{tabular}{|c|c|c|c|c|c|c|}
    \hline
    \multicolumn{7}{|c|}{\textbf{UAD TFR for ImageNet}}  \\
    \hhline{|=======|}
    {\footnotesize \revise{Tar / Src}} & R18 & R50 & D121 & Alex & VGG &Eff2s\\
    \hline
    R18 & \textbf{0.967} & 0.741 & 0.566 & 0.674 & 0.617 &0.603\\
    \hline
    R50 & 0.804 & \textbf{0.948} & 0.512 & 0.693 & 0.585 & 0.770\\
    \hline
    D121 & 0.545 & 0.571 & \textbf{0.900} & 0.454 & 0.549 & 0.710\\
    \hline
    Alex & 0.792 & 0.762 & 0.470 & \textbf{0.963} & 0.649 & 0.550\\
    \hline
    VGG & 0.698 & 0.667 & 0.518 & 0.674 & \textbf{0.946} & 0.554\\
    \hline
    Eff2s &0.795 & 0.785 & 0.760 & 0.452 & 0.503 & \textbf{0.952}\\
    \hhline{|=======|}
    \hline
    \multicolumn{7}{|c|}{\textbf{UAP TFR for ImageNet}}  \\
    \hhline{|=======|}
    {\footnotesize \revise{Tar / Src}} & R18 & R50 & D121 & Alex & VGG &Eff2s\\
    \hline
    R18 & \textbf{0.714} & 0.620 & 0.360 & 0.548 & 0.511 &0.423\\
    \hline
    R50 & 0.639 & \textbf{0.563} & 0.311 & 0.417 & 0.412 & 0.588\\
    \hline
    D121 & 0.318 & 0.352 & \textbf{0.679} & 0.313 & 0.350 & 0.423\\
    \hline
    Alex & 0.631 & 0.614 & 0.105 & \textbf{0.769} & 0.427 & 0.338\\
    \hline
    VGG & 0.540 & 0.484 & 0.222 & 0.540 & \textbf{0.798} & 0.252\\
    \hline
    Eff2s &0.439 & 0.529 & 0.523 & 0.213 & 0.241 & \textbf{0.788}\\
    \hhline{|=======|}
    \end{tabular}   
    \caption{Transferred fooling rates for UAD \& UAP on ImageNet.}
    \label{tab:tfr_im}
    \end{subtable}
    \caption{UAD and UAP cross-network transferability comparison.}
    \label{tab:UAD_cross}
\end{table}

\textbf{Transferability.} 
We further benchmarked the transferability (from the source DNN to other target DNNs) capabilities of UAD and UAP adversaries on ImageNet, TinyImageNet, CIFAR-100 and CIFAR-10. We compare UAD and UAP via Table \ref{tab:cosine_im}, which shows the cosine similarity scores between perturbations designed for different networks; and via Table \ref{tab:tfr_im}, which displays the accuracy-based transferred fooling rates (TFR in (\ref{Eq: TFR})) of perturbations when transferred from the source network (for which it was designed) and applied to attack the target network. 

As observed from Table \ref{tab:cosine_im}, while gradient-based UAPs designed for different DNNs were almost orthogonal to one another, the UADs achieve higher cosine similarity scores across the DNN architectures. Corresponding cosine similarity and TFR results for CIFAR-10, CIFAR-100 and TinyImageNet are included in Tables \ref{tab:cosine_c10}, \ref{tab:cosine_c100} \& \ref{tab:cosine_tim} and \ref{tab:tfr_c10}, \ref{tab:tfr_c100} \& \ref{tab:tfr_tim} of the Appendix. Finally, in Figures \ref{fig:SVDrops}, \ref{fig:SVDropstim}, \ref{fig:SVDropsc10} and \ref{fig:SVDropsc100} in the Appendix, we visualized the bar plot of the sorted singular values (descending order) for the attempted datasets and architectures. We observe that the loss's gradient matrix $G_S(f)$ for the UAD perturbation has always a unique top singular value.

\section{Conclusion}
In this work, we introduced universal adversarial directions (UADs) as a new variant of universal attacks. We provided theoretical evidence that while the universal adversarial example game lacks pure Nash equilibria, the universal adversarial direction game can possess an equilibrium with a pure strategy for the universal adversary. In addition, our numerical results indicate the improved transferability of the UAD adversary in comparison to gradient-based universal perturbations. Our analysis further introduces a potential extension of the UAD framework to rank-constrained adversarial attack and training schemes. Another interesting future direction to our work is to apply the proposed game-theoretic framework to analyze existing generative model-based adversarial perturbations.  {Furthermore, analyzing the challenging min-max optimization problem of UADs and their computational complexity is another potential extension of our proposed theoretical framework.}

\bibliographystyle{unsrt}
%

{
{
\bibliography{main}
%
}}

\begin{appendices}
\section{Proofs}
\subsection{Proof of Theorem \ref{Thm: UAP Nash}}
According to Theorem \ref{Thm: UAP Nash}'s assumption, for every perturbation $\boldsymbol{\delta}\in\mathbb{R}^d$, the following optimization problems are equivalent:
\begin{align*}\label{Eq: UAP Game}
     &\min_{f\in\mathcal{F}} \; \frac{1}{n}\sum_{i=1}^n\ell\bigl( f(\mathbf{x}_i+\boldsymbol{\delta}) , y_i \bigr) \\
     \stackrel{(a)}{\equiv} \; & \min_{f\in\mathcal{F},\, \mathbf{b}\in\mathbb{R}^d} \; \frac{1}{n}\sum_{i=1}^n\ell\bigl( f(\mathbf{x}_i+\mathbf{b}+\boldsymbol{\delta}) , y_i \bigr) \\ 
     \stackrel{(b)}{\equiv} \; & \min_{f\in\mathcal{F},\, \mathbf{b}'\in\mathbb{R}^d} \; \frac{1}{n}\sum_{i=1}^n\ell\bigl( f(\mathbf{x}_i+\mathbf{b}') , y_i \bigr)
\end{align*}
In the above, $(a)$ is a consequence of the theorem's assumption that for every function $f\in\mathcal{F}$ and bias vector $\mathbf{b}\in\mathbb{R}^d$, $f_{\mathbf{b}}\in\mathcal{F}$ is still a function in $\mathcal{F}$. Also, $(b)$ follows from the change of variable $\mathbf{b}' = \mathbf{b}+\boldsymbol{\delta}$ in the optimization problem. Since, the equivalent optimization problem has no dependence on perturbation $\boldsymbol{\delta}$, the optimal value of the optimization problem is independent from the choice of $\boldsymbol{\delta}$. Therefore, the proof of the theorem's first part is complete.

We give a proof by contradiction for the theorem's second part. To do this, we suppose that for a pure strategy $\boldsymbol{\delta}^*:\: \Vert \boldsymbol{\delta}^*\Vert \le \epsilon$ and a general mixed strategy over $\mathcal{F}$ characterized by probability distribution $P^*$, a Nash equilibrium in the universal adversarial example game is attained. The Nash equilibrium with pure strategy for the universal adversary implies that:
\begin{align*}
    \max_{\boldsymbol{\delta}:\: \Vert \boldsymbol{\delta}\Vert\le \epsilon}\; \mathbb{E}_{f\sim P^* }\biggl[\frac{1}{n}\sum_{i=1}^n\ell\bigl( f(\mathbf{x}_i+\boldsymbol{\delta}) , y_i \bigr)\biggr] \: &\le \, \mathbb{E}_{f\sim P^* }\biggl[\frac{1}{n}\sum_{i=1}^n\ell\bigl( f(\mathbf{x}_i+\boldsymbol{\delta}^*) , y_i \bigr)\biggr] \\
    \, &\le \, \min_{f\in\mathcal{F}}\: \frac{1}{n}\sum_{i=1}^n\ell\bigl( f(\mathbf{x}_i+\boldsymbol{\delta}^*) , y_i \bigr).
\end{align*}
However, as shown earlier, due to the theorem's assumption on function class $\mathcal{F}$, for every vector $\boldsymbol{\delta}\in\mathbb{R}^d$ we have
\begin{equation}
    \min_{f\in\mathcal{F}}\: \frac{1}{n}\sum_{i=1}^n\ell\bigl( f(\mathbf{x}_i+\boldsymbol{\delta}^*) , y_i \bigr) \; =  \; \min_{f\in\mathcal{F}}\: \frac{1}{n}\sum_{i=1}^n\ell\bigl( f(\mathbf{x}_i+\boldsymbol{\delta}) , y_i \bigr).
\end{equation}
Therefore, every $\widetilde{\boldsymbol{\delta}}\in\mathbb{R}^d$ satisfies:
\begin{align*}
     \max_{\boldsymbol{\delta}:\: \Vert \boldsymbol{\delta}\Vert\le \epsilon}\; \mathbb{E}_{f\sim P^* }\biggl[\frac{1}{n}\sum_{i=1}^n\ell\bigl( f(\mathbf{x}_i+\boldsymbol{\delta}) , y_i \bigr)\biggr] \, &\le \, \min_{f\in\mathcal{F}}\: \frac{1}{n}\sum_{i=1}^n\ell\bigl( f(\mathbf{x}_i+\widetilde{\boldsymbol{\delta}}) , y_i \bigr)\\
     &\le\, \mathbb{E}_{f\sim P^* }\biggl[\frac{1}{n}\sum_{i=1}^n\ell\bigl( f(\mathbf{x}_i+\widetilde{\boldsymbol{\delta}}) , y_i \bigr)\biggr].
\end{align*}
The above inequalities imply that the function $g^*(\boldsymbol{\delta}):= \mathbb{E}_{f\sim P^* }\bigl[\frac{1}{n}\sum_{i=1}^n\ell\bigl( f(\mathbf{x}_i+\boldsymbol{\delta}) , y_i \bigr)\bigr]$ takes the same minimum value for every $\epsilon$-norm-bounded $\boldsymbol{\delta}$. As a result, the existence of a Nash equilibrium with a pure adversary strategy implies that at the optimal classifier strategy  every norm-bounded universal perturbation including the zero perturbation leads to a Nash equilibrium, and the minimum averaged loss does not change by adding any non-zero perturbations. This contradiction of attaining a Nash equilibrium with a trivial zero-universal-perturbation completes the theorem's proof. 

\subsection{Proof of Proposition \ref{Prop: PCA approximation}}
To show this proposition, note that under the smoothness assumption in the paper, we have:
\begin{align*}
  &\ell\bigl(f(\mathbf{x}_i ),y_i \bigr) + \tau_i\delta^\top\nabla_{\mathbf{x}}\ell\bigl(f(\mathbf{x}_i ),y_i \bigr) - \frac{\rho}{2} \tau_i^2 \Vert \boldsymbol{\delta}\Vert^2 \\ 
    \le\; &\ell\bigl(f(\mathbf{x}_i + \tau_i\boldsymbol{\delta}),y_i \bigr) \numberthis \\
    \le\; &\ell\bigl(f(\mathbf{x}_i ),y_i \bigr) + \tau_i\delta^\top\nabla_{\mathbf{x}}\ell\bigl(f(\mathbf{x}_i ),y_i \bigr)+\frac{\rho}{2} \tau_i^2 \Vert \boldsymbol{\delta}\Vert^2.
\end{align*}
As a result, since $\Vert \boldsymbol{\delta}\Vert^2\le B$, we obtain the followings:
\begin{align}
   &\max_{\tau_i\in\mathbb{R} }\: \biggl\{ \ell\bigl(f(\mathbf{x}_i ),y_i \bigr) + \tau_i\delta^\top\nabla_{\mathbf{x}}\ell\bigl(f(\mathbf{x}_i ),y_i \bigr)- \frac{\lambda+B\rho}{2} \tau_i^2\biggr\} \\
    \le\;&\max_{\tau_i\in\mathbb{R} }\:\biggl\{\ell\bigl(f(\mathbf{x}_i + \tau_i\boldsymbol{\delta}),y_i \bigr)-\frac{\lambda}{2} \tau_i^2\biggr\}  \\
    \le \; &\max_{\tau_i\in\mathbb{R} }\:\biggl\{\ell\bigl(f(\mathbf{x}_i ),y_i \bigr) + \tau_i\delta^\top\nabla_{\mathbf{x}}\ell\bigl(f(\mathbf{x}_i ),y_i \bigr) -\frac{\lambda-B\rho}{2} \tau_i^2 \biggr\}.
\end{align}
Note that both the upper-bound and lower-bound in the above inequalities represent quadratic optimization problems, where under the assumption that $\lambda> B\rho$, the optimal solutions to the lower-bound and upper-bound optimization problems will be the followings implied by the first-order necessary condition:
\begin{equation}
    \tau_i^l = \frac{1}{\lambda+B\rho} \delta^\top\nabla_{\mathbf{x}}\ell\bigl(f(\mathbf{x}_i ),y_i \bigr),\quad \tau_i^u = \frac{1}{\lambda-B\rho} \delta^\top\nabla_{\mathbf{x}}\ell\bigl(f(\mathbf{x}_i ),y_i \bigr).
\end{equation}
Plugging the optimal solutions into the bounds will lead to the following inequalities:
\begin{align*}
   &\ell\bigl(f(\mathbf{x}_i ),y_i \bigr) + \frac{1}{2(\lambda+B\rho)} \left( \delta^\top\nabla_{\mathbf{x}}\ell\bigl(f(\mathbf{x}_i ),y_i \bigr)\right)^2 \\
    \le\; &\max_{\tau_i\in\mathbb{R} }\biggl\{\ell\bigl(f(\mathbf{x}_i + \tau_i\boldsymbol{\delta}),y_i \bigr)-\frac{\lambda}{2} \tau_i^2\biggr\}  \numberthis \\
    \le\;  &\ell\bigl(f(\mathbf{x}_i ),y_i \bigr) + \frac{1}{2(\lambda-B\rho)} \left( \delta^\top\nabla_{\mathbf{x}}\ell\bigl(f(\mathbf{x}_i ),y_i \bigr)\right)^2.
\end{align*}
Therefore, the proof is complete.
\subsection{Proof of Theorem \ref{Thm: UAD Nash}}
Consider the target max-min optimization problem:
\begin{align*}
\max_{\Vert\boldsymbol{\delta}\Vert \le 1}\, \min_{f\in\mathcal{F}} \, \frac{1}{n}\sum_{i=1}^n\biggl[ \ell\bigl(f(\mathbf{x}_i),y_i \bigr) + \eta \bigl(\boldsymbol{\delta}^\top\nabla_{\mathbf{x}}\ell\bigl(f(\mathbf{x}_i ),y_i\bigr)\bigr)^2\biggr]  = L_S(f) + \eta \boldsymbol{\delta}^\top G_S(f)G_S(f)^\top\boldsymbol{\delta}.
\end{align*}
In the above max-min optimization problem, we can use the trace operator to rewrite the objective function as
\begin{align*}
   &L_S(f) + \eta \boldsymbol{\delta}^\top G_S(f)G_S(f)^\top\boldsymbol{\delta} \\
   = \; &L_S(f) + \eta \operatorname{Tr}\bigl(\boldsymbol{\delta}^\top G_S(f)G_S(f)^\top\boldsymbol{\delta} \bigr) \\
   =\; & L_S(f) + \eta \operatorname{Tr}\bigl( G_S(f)G_S(f)^\top\boldsymbol{\delta}\boldsymbol{\delta}^\top \bigr)
\end{align*}
Here $\operatorname{Tr}(\cdot)$ denotes the trace operator, and the above equality holds since the trace operator is linear and satisfies $\operatorname{Tr}(AB)=\operatorname{Tr}(BA)$ as long as the matrix multiplications $AB,\,BA$ are well-defined. Based on the above discussion, we apply a change of variables and define the matrix variable $\Delta = \boldsymbol{\delta}\boldsymbol{\delta}^\top$ which leads to the following equivalent max-min optimization problem: 
\begin{align*}
\max_{\substack{\Delta\in \mathcal{S}^d_+:\: \Vert\Delta\Vert_* \le 1 \\
   \operatorname{rank}(\Delta)\le 1}}\, \min_{f\in\mathcal{F}} \,  L_S(f) + \eta \operatorname{Tr}\bigl( G_S(f)G_S(f)^\top\Delta \bigr).
\end{align*}
Here, $\mathcal{S}^d_+$ denotes the $d\times d$-positive-semidefinite (PSD) cone, $\Vert\cdot\Vert_*$ stands for the nuclear norm, i.e. the sum of a matrix's singular values, and $\operatorname{rank}(\cdot)$ is the rank of a matrix. Note that the constraints on matrix variable $\Delta$ requires $\Delta = \boldsymbol{\delta}\boldsymbol{\delta}^\top$ for some vector $\Vert \boldsymbol{\delta}\Vert\le 1$. Also, since $G_S(f)G_S(f)^\top$ and $\Delta$ are both PSD matrices, every solution to the above bilevel optimization problem will take the maximum allowable norm value, i.e. $\Vert{\Delta}^*\Vert_* = 1$ or equivalently for a PSD matrix we have  $\operatorname{Tr}({\Delta}^*) = 1$. Therefore, assuming the loss function only takes non-negative values, the above max-min problem has the same solution as the following max-min problem
\begin{align*}
&\max_{\substack{\Delta\in \mathcal{S}^d_+:\: \Vert\Delta\Vert_* \le 1 \\
   \operatorname{rank}(\Delta)\le 1}}\: \min_{f\in\mathcal{F}} \:  L_S(f)\operatorname{Tr}(\Delta) + \eta \operatorname{Tr}\bigl( G_S(f)G_S(f)^\top\Delta \bigr) \\
   \, =\; &\max_{\substack{\Delta\in \mathcal{S}^d_+:\: \Vert\Delta\Vert_* \le 1 \\
   \operatorname{rank}(\Delta)\le 1}}\: \min_{f\in\mathcal{F}} \: \operatorname{Tr}\biggl( \bigl(L_S(f)I_d + \eta G_S(f)G_S(f)^\top \bigr) \Delta \biggr) . \numberthis
\end{align*}
We can define the equivalent problem using the matrix set $\mathcal{M_F}:= \bigl\{L_S(f)I_d + \eta G_S(f)G_S(f)^\top :\: f\in\mathcal{F} \bigr\}$:  
\begin{align}\label{Proof: Theorem 2, equation 5}
\max_{\substack{\Delta\in \mathcal{S}^d_+:\: \Vert\Delta\Vert_* \le 1 \\
   \operatorname{rank}(\Delta)\le 1}}\, \min_{M\in\mathcal{M_F}} \,   \operatorname{Tr}\bigl( M^\top \Delta \bigr) . \numberthis
\end{align}
According to the theorem's assumption $\mathcal{M_F}$ is a convex and compact subset of PSD matrices. Also, note that the objective function $\operatorname{Tr}\bigl( M^\top \Delta \bigr)$ is bi-linear in PSD matrix variables $\Delta$ and $M$. We also note that the following superset of the maximization problem's feasible set $\{\Delta\in \mathcal{S}^d_+:\: \Vert\Delta\Vert_* \le 1\}$ is by definition convex and compact. As a result, Sion's minimax theorem \cite{sion1958general} implies that the following min-max and max-min problems share a common saddle-point solution $(\Delta^*,M^*)$:
\begin{equation}\label{Proof: Theorem 2, equation 4}
\max_{\substack{\Delta\in \mathcal{S}^d_+:\: \Vert\Delta\Vert_* \le 1}}\, \min_{M\in\mathcal{M_F}} \,   \operatorname{Tr}\bigl( M^\top \Delta \bigr) \; = \;  \min_{M\in\mathcal{M_F}} \, \max_{\substack{\Delta\in \mathcal{S}^d_+:\: \Vert\Delta\Vert_* \le 1}}\,  \operatorname{Tr}\bigl( M^\top \Delta \bigr) . \numberthis
\end{equation}
However, note that since the $L_2$-operator norm ($\Vert\cdot\Vert_2$) and nuclear norms are dual to each other:
\begin{align*}
\max_{\substack{\Delta\in \mathcal{S}^d_+:\: \Vert\Delta\Vert_* \le 1}}\,  \operatorname{Tr}\bigl( M^\top \Delta \bigr) \; = \; \Vert M \Vert_2.
\end{align*}
Therefore, based on the theorem's first assumption that for every minimizer $f^*\in \mathcal{F}$ for the min-max problem, the matrix $G_S(f^*)$ has a unique top singular value or equivalently the PSD matrix $M_{f^*} = L_S(f^*)I_d + \eta G_S(f^*)G_S(f^*)^\top$ has a unique top eigenvalue, then the corresponding maximization solution $\Delta^*_{f^*}$ will be rank-1, as the matrix's nuclear norm needs to be concentrated on the top right-singular vector of $G_S(f^*)$. As a result, there exists a shared solution $(f^*,\Delta^*)$ for the min-max and max-min problems in \eqref{Proof: Theorem 2, equation 4} where $\Delta^*$ is a rank-1 matrix. Since this solution satisfies the maximization constraints of the original max-min problem in \eqref{Proof: Theorem 2, equation 5} with the maximization feasible set being a subset of the feasible set in \eqref{Proof: Theorem 2, equation 5}, $(f^*,\Delta^*)$ will also be a solution to \eqref{Proof: Theorem 2, equation 5}. Similarly, $(f^*,\Delta^*)$ is also a solution to the min-max version of \eqref{Proof: Theorem 2, equation 5}, since the max-min inequality implies that
\begin{align}
\max_{\substack{\Delta\in \mathcal{S}^d_+:\: \Vert\Delta\Vert_* \le 1 \\
   \operatorname{rank}(\Delta)\le 1}}\, \min_{M\in\mathcal{M_F}} \,   \operatorname{Tr}\bigl( M^\top \Delta \bigr) \; \le \;  \min_{M\in\mathcal{M_F}} \, \max_{\substack{\Delta\in \mathcal{S}^d_+:\: \Vert\Delta\Vert_* \le 1 \\
   \operatorname{rank}(\Delta)\le 1}}\,  \operatorname{Tr}\bigl( M^\top \Delta \bigr)
\end{align}
and in the above min-max problem, matrix $M_{f^*}$ achieves the lower-bound given by the max-min formulation. Thus, $(f^*,\Delta^*)$ is a saddle-point for the original max-min optimization problem, and therefore results in a pure Nash equilibrium to the approximate universal adversarial direction game. Hence, the proof of the theorem's first part is finished.

Next, under the theorem's second assumption that for every minimizer $f^*\in \mathcal{F}$ of the min-max problem, the corresponding matrix $G_S(f^*)$ has a top singular value with multiplicity at most $r$ or equivalently the PSD matrix $M_{f^*} = L_S(f^*)I_d + \eta G_S(f^*)G_S(f^*)^\top$ has a maximum eigenvalue with multiplicity at most $r$, then we have a saddle point solution $(\Delta^*,M_{f^*})$ for \eqref{Proof: Theorem 2, equation 4} where $\Delta^*$ is of rank $r$. Therefore, we assume that the orthonormal unit-norm vectors in $\{\boldsymbol{\delta}^*_1,\ldots,\boldsymbol{\delta}^*_r \}$ are the top eigenvectors of  $\Delta^*$. Note that the solution $(\Delta^*,M_{f^*})$ will solve the following problem since the maximization problem's feasible set in the following problem is a subset of the the one in \eqref{Proof: Theorem 2, equation 4}.
\begin{align}
\max_{\substack{\Delta\in \mathcal{S}^d_+:\: \Vert\Delta\Vert_* \le 1 \\
   \operatorname{rank}(\Delta)\le r}}\, \min_{M\in\mathcal{M_F}} \,   \operatorname{Tr}\bigl( M^\top \Delta \bigr) .
\end{align}
In addition, $(\Delta^*,M_{f^*})$ will solve the min-max problem corresponding to the above task, because it achieves the lower-bound coming from the following max-min inequality
\begin{align}
\max_{\substack{\Delta\in \mathcal{S}^d_+:\: \Vert\Delta\Vert_* \le 1 \\
   \operatorname{rank}(\Delta)\le r}}\, \min_{M\in\mathcal{M_F}} \,   \operatorname{Tr}\bigl( M^\top \Delta \bigr) \; \le \; \min_{M\in\mathcal{M_F}} \, \max_{\substack{\Delta\in \mathcal{S}^d_+:\: \Vert\Delta\Vert_* \le 1 \\
   \operatorname{rank}(\Delta)\le r}}\,    \operatorname{Tr}\bigl( M^\top \Delta \bigr).
\end{align}
As a result, the rank-$r$ $\Delta^*$ combined with a mixed strategy for the classifier player choosing $f\in\mathcal{F}$  results in a mixed Nash equilibrium for the following max-min game:
\begin{align}
\max_{\substack{\Delta\in \mathcal{S}^d_+:\: \Vert\Delta\Vert_* \le 1 \\
   \operatorname{rank}(\Delta)\le 1}}\, \min_{P_f\in\mathcal{P_F}} \,  \mathbb{E}_{f\sim P_f}\biggl[ L_S(f)\operatorname{Tr}(\Delta) + \eta \operatorname{Tr}\bigl( G_S(f)G_S(f)^\top\Delta \bigr) \biggr].
\end{align}
This Nash equilibrium implies the existence of mixed strategies for the universal adversarial direction and classifier players where the universal adversary always generates the perturbation from the rank-$r$ subspace of $\Delta^*$'s range spanned by orthonormal vectors in $\{\boldsymbol{\delta}^*_1,\ldots,\boldsymbol{\delta}^*_r \}$. Therefore, the theorem's proof is complete.
\section{Additional Numerical Experiments}
\subsection{Additional Numerical Results on UAD-PCA, UAD-grad, and UAP-grad Attacks}
\begin{figure}[H]
\centering
  \includegraphics[width=1.0\textwidth]{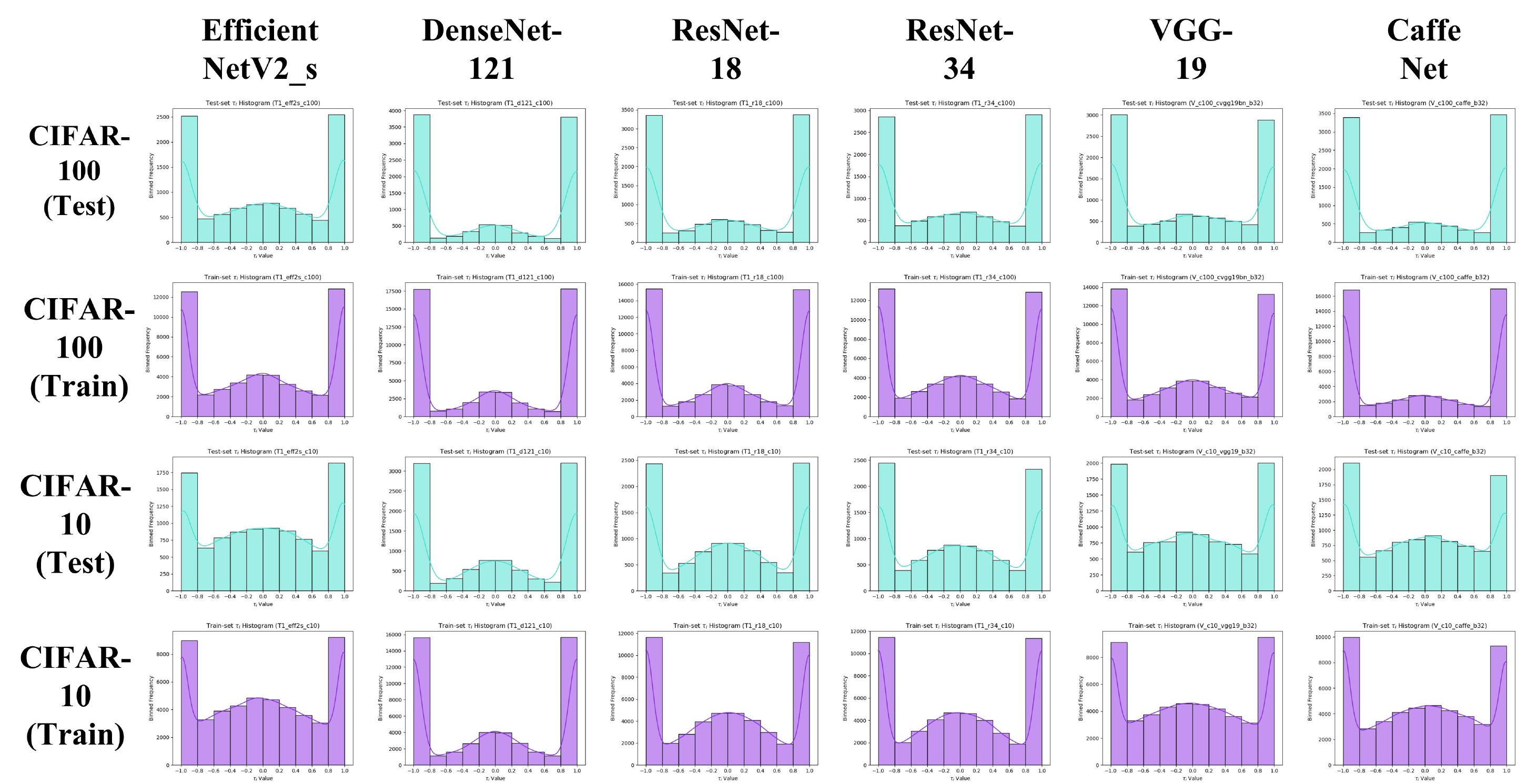}
  \caption{{Distribution of $\tau_i \in [-1, 1]$ of UAD attacks on CIFAR-100 and CIFAR-10.}}
  \label{fig: tau1}
\end{figure}

\begin{figure}[H]
\centering
  \includegraphics[width=1.0\textwidth]{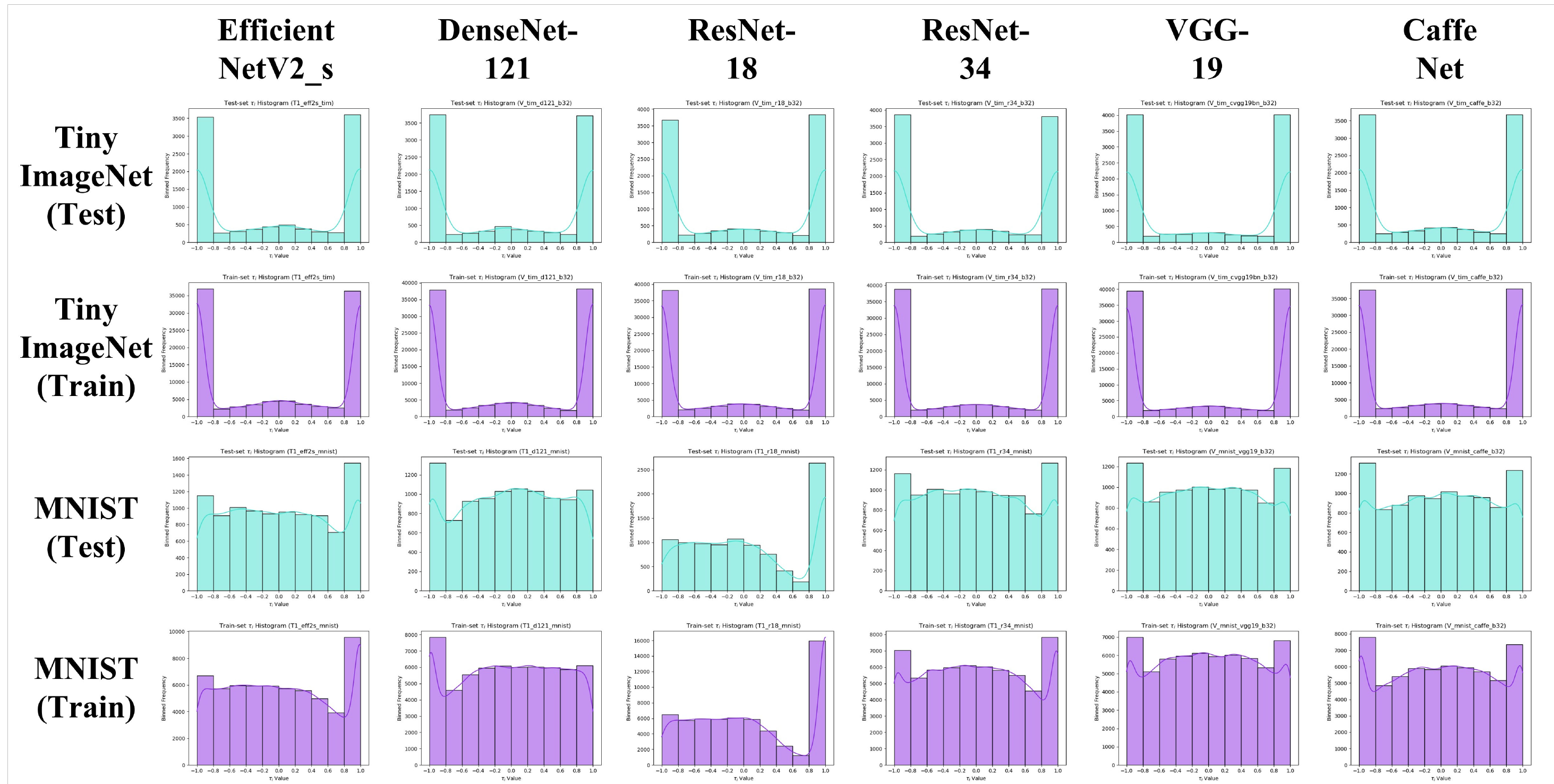}
  \caption{{Distribution of $\tau_i \in [-1, 1]$ of UAD attacks on TinyImageNet and MNIST.}}
  \label{fig: tau2}
\end{figure}

\begin{figure}[h]
\centering
  \includegraphics[width=0.99\textwidth]{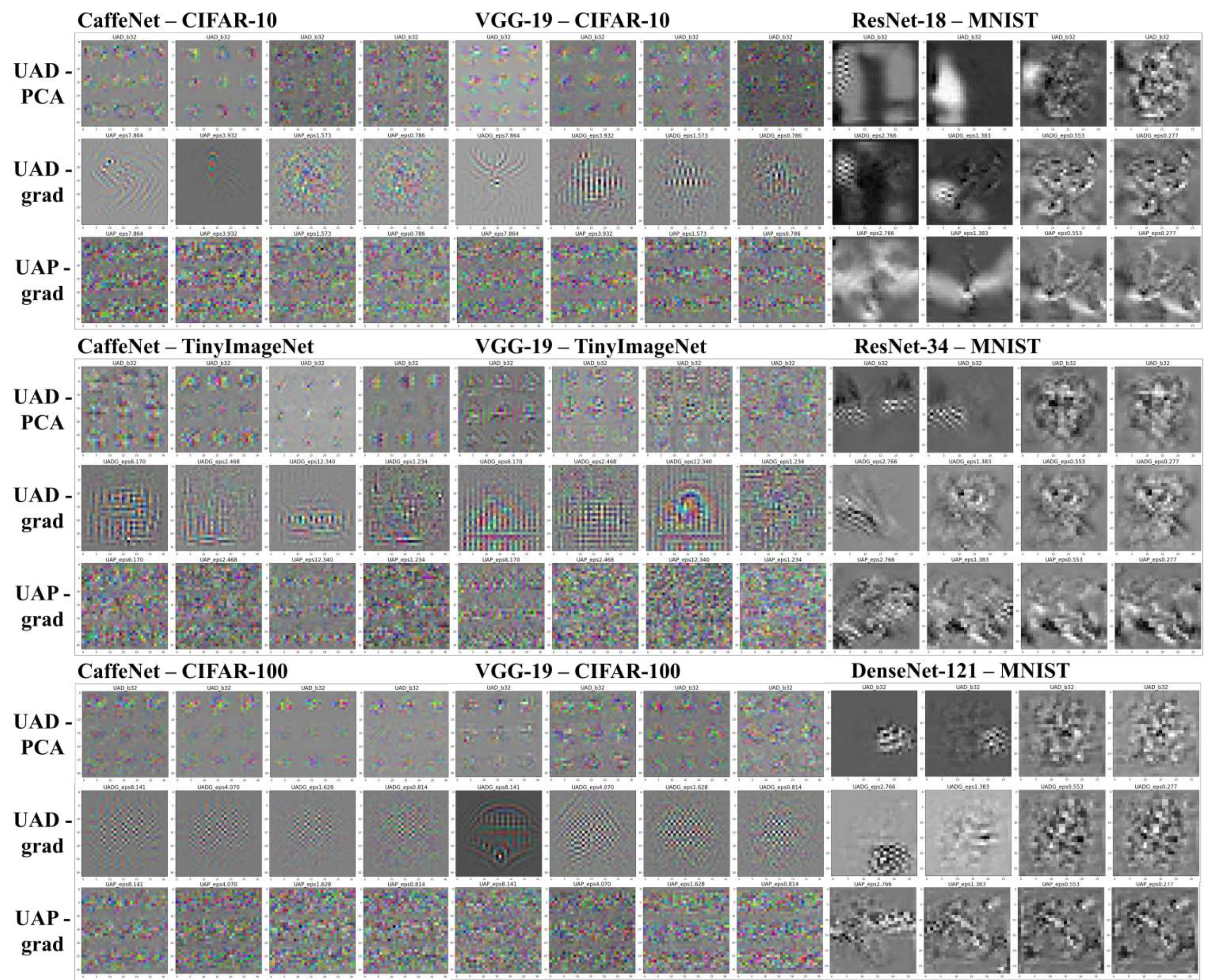}
  \caption{Visualizations of UAD-PCA, UAD-grad, UAP-grad adversarial noise at $\epsilon=0.1, 0.05, 0.02, 0.01 \cdot \mathbb{E}[\Vert \textbf{X} \Vert_2]$}
  \label{fig: visuals2}
\end{figure}

Here, we present the complete numerical results for the visualizations of adversarial perturbations generated by UAD-PCA, UAD-grad and UAP-grad in Figure \ref{fig: visuals2}. The proposed UAD-PCA and UAD-grad both generated rather semantically meaningful noise patterns, while UAP-grad synthesizes seemingly less meaningful perturbations without a noteworthy pattern. We further report the fooling rates and achieved adversarial test accuracies of the three universal attack algorithms, on CIFAR-10, CIFAR-100 and TinyImageNet datasets. Note that the adversarial test accuracies are presented in Table \ref{tab:strength2}; fooling rates are given in Tables \ref{tab:fool_UAD}, \ref{tab:fool_UAP}, \ref{tab:fool_UADG} {and \ref{tab:fool_GAP}}. {Furthermore, in histogram Figures \ref{fig: tau1}, \ref{fig: tau2}, we visualize the distribution of optimal $\tau_i$'s for UAD attacks on various datasets and architectures (including EfficientNetV2-S). We can see that the UAD framework is more expressive than UAP, allowing for both positive and negative $\tau_i$ perturbation coefficients with varying strengths between $[-1, 1]$.} 

\subsection{Additional Numerical Results on Generalizability and Transferability of UADs vs. UAPs}
To measure the transferability and strength of the perturbations, we quantified the explained transferred fooling rate of UADs and UAPs across different DNN architectures. $\tau_i \boldsymbol{\delta}_a$ is the transferred (designed for a different model) universal attack's perturbation for sample $\boldsymbol{x}_i$, while $\tau_i \boldsymbol{\delta}_m$ is the original (designed for the same model) attack's perturbation. Consider the following evaluation measures:
\begin{align*}
    TP = \sum_{i} \left(f(\boldsymbol{x}_i +\tau_i \boldsymbol{\delta}_a) \neq f(\boldsymbol{x}_i)\right) \: &\&\: \left(f(\boldsymbol{x}_i +\tau_i \boldsymbol{\delta}_a) = f(\boldsymbol{x}_i +\tau_i \boldsymbol{\delta}_m)\right)\\
    TN = \sum_{i} \left(f(\boldsymbol{x}_i +\tau_i \boldsymbol{\delta}_a) \neq f(\boldsymbol{x}_i)\right) \: &\&\: \left(f(\boldsymbol{x}_i +\tau_i \boldsymbol{\delta}_a) \neq f(\boldsymbol{x}_i +\tau_i \boldsymbol{\delta}_m)\right)\\
    FP = \sum_{i} \left(f(\boldsymbol{x}_i +\tau_i \boldsymbol{\delta}_a) = f(\boldsymbol{x}_i)\right) \: &\&\: \left(f(\boldsymbol{x}_i +\tau_i \boldsymbol{\delta}_a) \neq f(\boldsymbol{x}_i +\tau_i \boldsymbol{\delta}_m)\right)\\
    FN = \sum_{i} \left(f(\boldsymbol{x}_i +\tau_i \boldsymbol{\delta}_a) = f(\boldsymbol{x}_i)\right) \: &\&\: \left(f(\boldsymbol{x}_i +\tau_i \boldsymbol{\delta}_a) \neq f(\boldsymbol{x}_i +\tau_i \boldsymbol{\delta}_m)\right)
\end{align*}   
We defined and evaluated the following \emph{Transferred Fooling Rate (TFR)} score in our experiments:
\begin{align}\label{Eq: TFR}
    \textrm{TFR} &= \frac{TP+TN}{TP + FN + TN + FP}
\end{align}
Intuitively, TP (true positive / explained and transferred ability to fool) means both the transferred and original adversarial perturbations fool the source model (i.e., the model prediction on the transfer-perturbed sample is not equal to that on unperturbed sample but is equal to that on the original-perturbed sample); TN (true negative / unexplained and untransferred ability to fool) means the unexplained ability to fool the source model; FP (false positive / explained and transferred inability to fool) means neither the original nor transferred adversarial perturbations are able to fool the network; FN (false negative / unexplained and untransferred inability to fool) means the transferred attack neither fools the model nor does it have a prediction that corresponds to the original perturbed result. Since the measurement already accounts for the transferability between models, a higher TFR indicates both better transferability and fooling ability. We measured TFR scores across different DNN architectures and datasets; the results suggest that our proposed UAD attack exhibits higher transferability than gradient-based UAPs.

\begin{figure}[h!]
\centering
  \includegraphics[width=\textwidth]{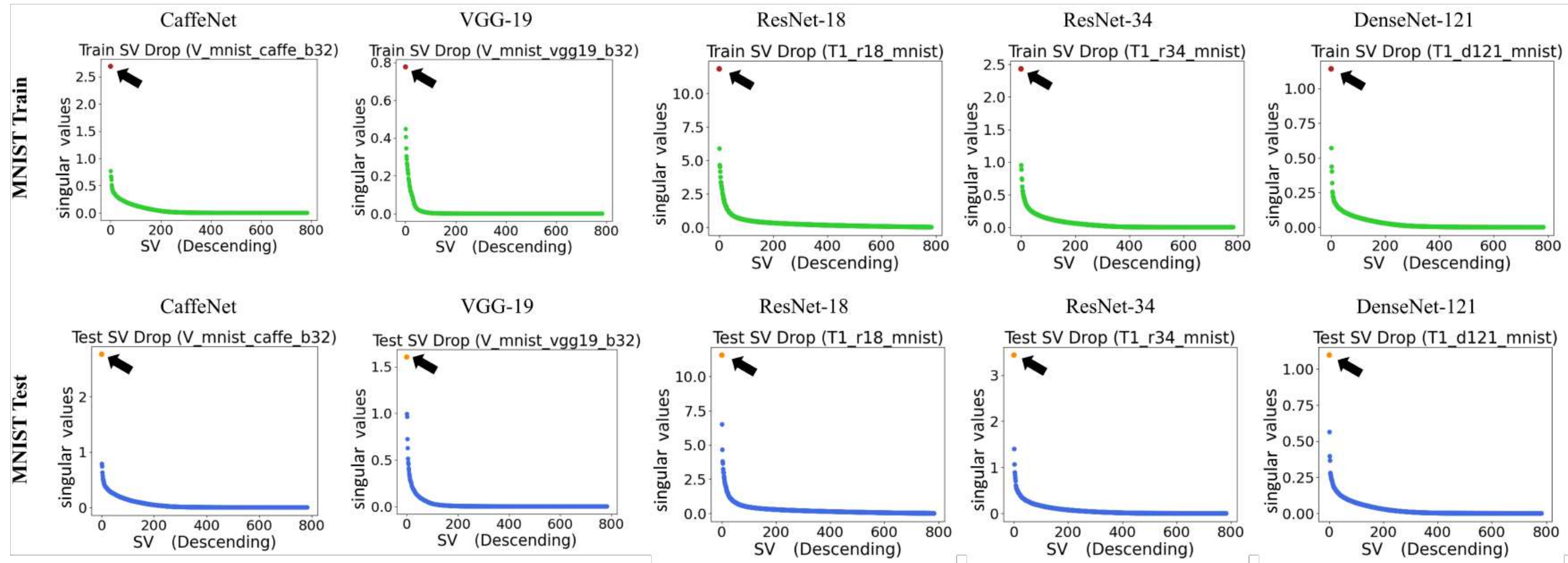}
  \caption{Singular values of $G_S(f)$ on MNIST; top singular value is denoted with an arrow.}
  \label{fig:SVDrops}
\end{figure}

\begin{figure}[h!]
\centering
  \includegraphics[width=\textwidth]{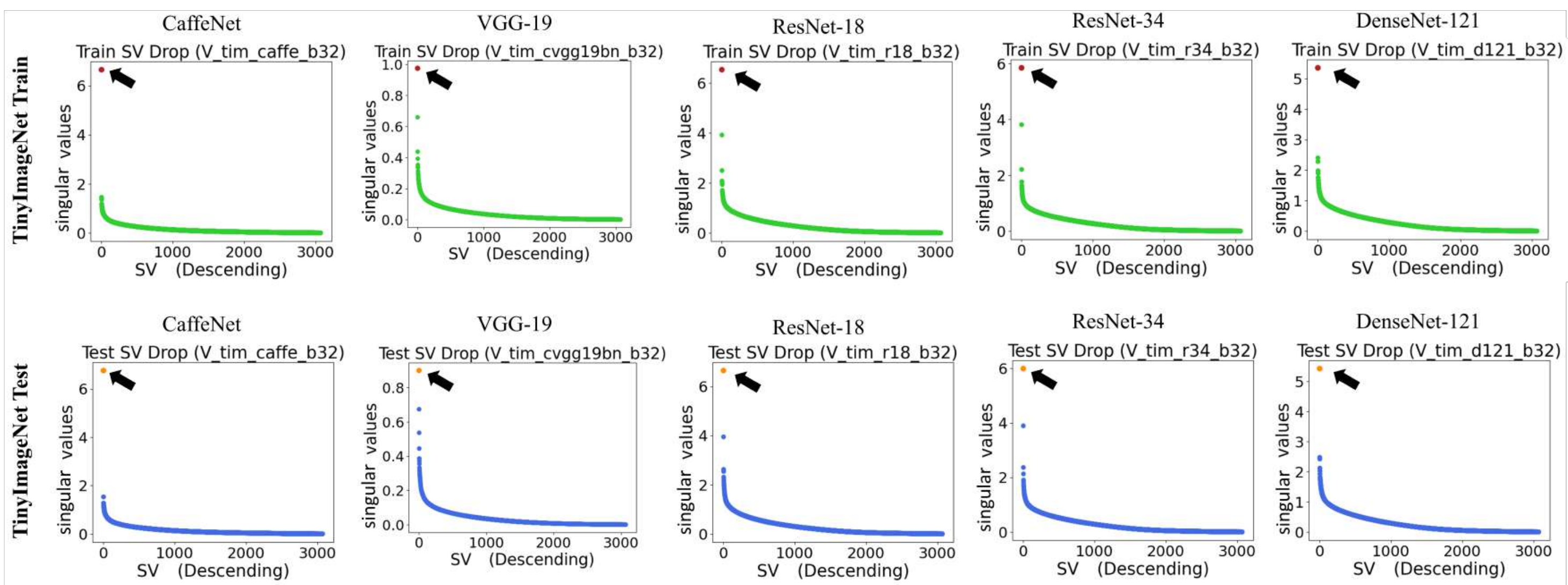}
  \caption{Singular values of $G_S(f)$ on TinyImageNet train and test sets; the top singular value is denoted with an arrow.}
  \label{fig:SVDropstim}
\end{figure}

\begin{figure}[h!]
\centering
  \includegraphics[width=\textwidth]{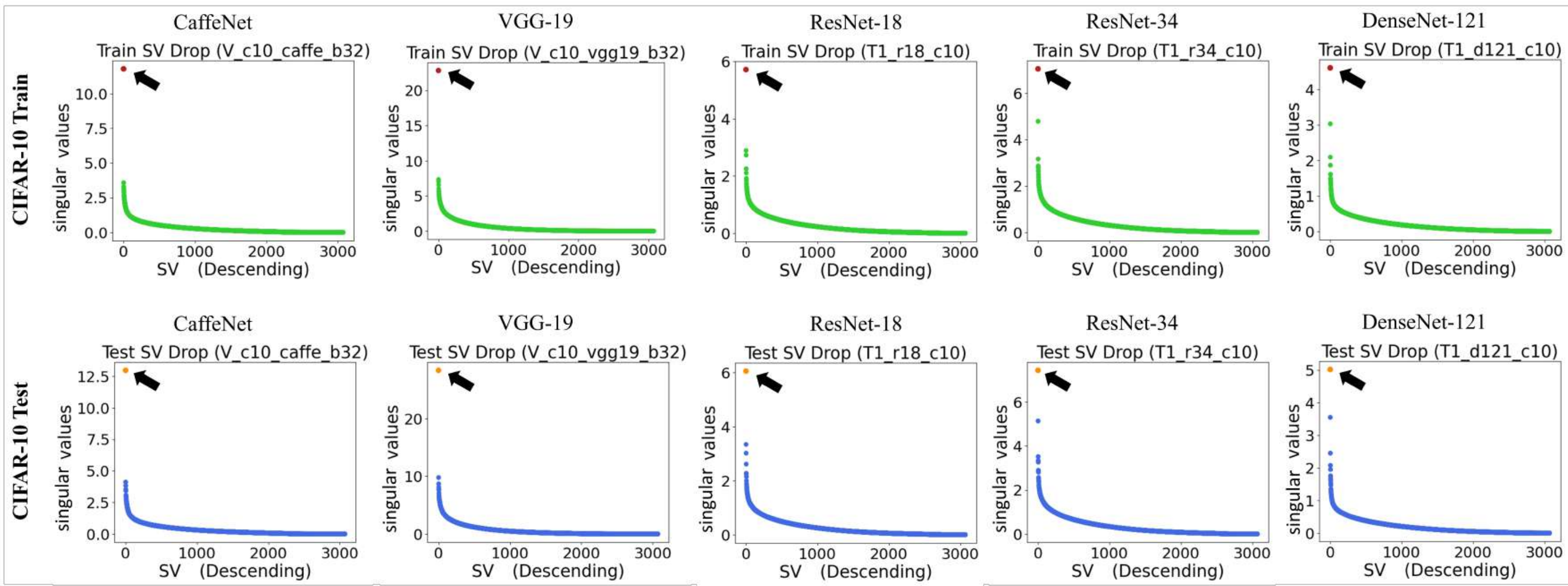}
  \caption{Singular values of $G_S(f)$ on CIFAR-10 train and test sets; the top singular value is denoted with an arrow.}
  \label{fig:SVDropsc10}
\end{figure}

\begin{figure}[h!]
\centering
\includegraphics[width=\textwidth]{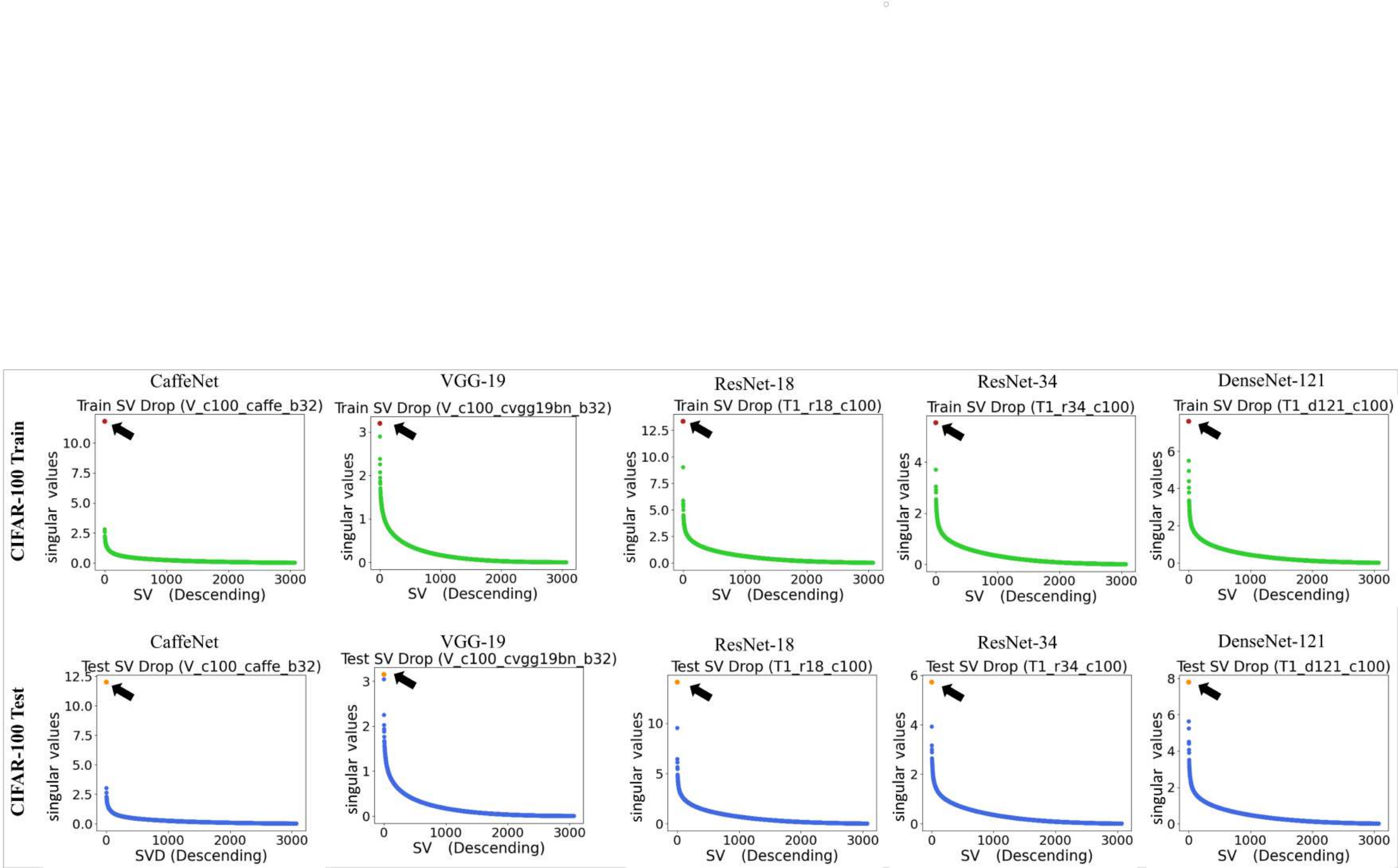}
  \caption{Singular values of $G_S(f)$ on CIFAR-100 train and test sets; the top singular value is denoted with an arrow.}
  \label{fig:SVDropsc100}
\end{figure}

Finally, singular value decomposition (SVD) was performed on the matrix of the loss function's gradients with respect to the input image samples on the TinyImageNet dataset, in order to compare the universality of perturbation. These results are visualized in Figures \ref{fig:SVDrops}, \ref{fig:SVDropstim}, \ref{fig:SVDropsc10}, \ref{fig:SVDropsc100}. We observed that the loss's gradient matrix $G_S(f)$ used for generating the UAD perturbations has always a unique top singular value across train and test sets.

\captionsetup[figure]{font=normalsize}
\begin{table}[H]
    \renewcommand{\arraystretch}{1.1}
    \centering
    \footnotesize
    \resizebox{0.85\linewidth}{!}{
    \begin{tabular}{|c|c|c|c|c|c|c|c|}
    \cline{1-8}
    \hline
    \multicolumn{8}{|c|}{\textbf{UAD-PCA Fooling Rates}}  \\
    \hhline{|========|}
    \multicolumn{2}{|c|}{\backslashbox{\textbf{Dataset}}{\textbf{Model}}} & ResNet-18 & ResNet-34 & DenseNet-121 & CaffeNet & VGG-19 & \revise{EfficientNet}\\
         \hline
    \multirow{4}{*}{CIFAR-10 (Train)} &$\epsilon=0.1$  &87.930 &85.948  &87.230  &73.874  &70.704 &\revise{71.670}\\
                                      &$\epsilon=0.05$ &49.860 &57.962  &77.420  &25.788  &25.878 &\revise{28.062}\\
                                      &$\epsilon=0.02$ &6.574  &7.286   &28.374  &10.380  &8.544 &\revise{9.818}\\
                                      &$\epsilon=0.01$ &2.204  &3.304   &9.338   &9.658   &5.268 &\revise{9.170}\\
    \hline
    \multirow{4}{*}{CIFAR-10 (Test)} &$\epsilon=0.1$  &88.350 &85.380  &86.940  &74.010  &71.180 &\revise{69.410}\\
                                     &$\epsilon=0.05$ &51.110 &57.500  &77.700  &24.050  &24.720 &\revise{23.240}\\
                                     &$\epsilon=0.02$ &6.480  &8.560   &30.340  &4.130   &8.070 &\revise{2.910}\\
                                     &$\epsilon=0.01$ &2.080  &2.380   &10.520  &1.350   &1.960 &\revise{1.180}\\
    \hhline{|========|}
    \multirow{4}{*}{CIFAR-100 (Train)} &$\epsilon=0.1$  &91.604  &89.034  &97.326  &95.926  &86.738 &\revise{88.414}\\
                                       &$\epsilon=0.05$ &70.064  &61.936  &90.630  &77.440  &68.436 &\revise{57.102}\\
                                       &$\epsilon=0.02$ &33.416  &24.396  &45.984  &40.958  &29.776 &\revise{28.056}\\
                                       &$\epsilon=0.01$ &13.368  &9.822   &15.164  &30.364  &14.846 &\revise{21.326}\\
    \hline
    \multirow{4}{*}{CIFAR-100 (Test)} &$\epsilon=0.1$  &91.910  &89.330  &97.540  &95.300  &86.980 &\revise{84.840}\\
                                      &$\epsilon=0.05$ &72.130  &65.050  &91.740  &73.990  &71.090 &\revise{49.000}\\
                                      &$\epsilon=0.02$ &40.010  &31.130  &52.460  &30.550  &34.970 &\revise{14.800}\\
                                      &$\epsilon=0.01$ &18.650  &13.040  &23.730  &11.070  &12.620 &\revise{2.980}\\
    \hhline{|========|}
    \multirow{4}{*}{TinyImageNet (Train)} &$\epsilon=0.1$  &98.101  &98.358  &98.096  &98.368  &99.023 &\revise{98.390}\\
                                          &$\epsilon=0.05$ &83.353  &82.876  &79.080  &80.425  &93.549 &\revise{88.526}\\
                                          &$\epsilon=0.02$ &62.006  &64.609  &59.091  &70.242  &71.471 &\revise{69.781}\\
                                          &$\epsilon=0.01$ &56.095  &53.904  &48.460  &66.078  &67.022 &\revise{62.069}\\
    \hline
    \multirow{4}{*}{TinyImageNet (Test)} &$\epsilon=0.1$  &98.160  &98.550  &98.260  &98.100  &99.400 &\revise{97.380}\\
                                         &$\epsilon=0.05$ &77.360  &77.660  &74.160  &60.410  &91.950 &\revise{76.700}\\
                                         &$\epsilon=0.02$ &40.120  &48.740  &41.280  &31.160  &42.780 &\revise{33.790}\\
                                         &$\epsilon=0.01$ &25.920  &20.110  &19.910  &13.680  &19.640 &\revise{9.580}\\
    \hline
    \end{tabular}}
    \vspace{2mm}
    \caption{UAD-PCA fooling rates with $\epsilon=0.1, 0.05, 0.02, 0.01 \cdot \mathbb{E}[\Vert \mathbf{X}\Vert_2]$.}
    \label{tab:fool_UAD}
\end{table}

\captionsetup[figure]{font=normalsize}
\begin{table}[H]
    \renewcommand{\arraystretch}{1.1}
    \centering
    \footnotesize
    \resizebox{0.85\linewidth}{!}{
    \begin{tabular}{|c|c|c|c|c|c|c|c|}
    \cline{1-8}
    \hline
    \multicolumn{8}{|c|}{\textbf{UAD-grad Fooling Rates}}  \\
    \hhline{|========|}
    \multicolumn{2}{|c|}{\backslashbox{\textbf{Dataset}}{\textbf{Model}}} & ResNet-18 & ResNet-34 & DenseNet-121 & CaffeNet & VGG-19 & \revise{EfficientNet}\\
         \hline
    \multirow{4}{*}{CIFAR-10 (Train)} &$\epsilon=0.1$  &79.820 &69.328  &85.806  &67.570  &44.644 &\revise{67.994}\\
                                      &$\epsilon=0.05$ &15.372 &18.410  &70.008  &13.824  &14.610 &\revise{16.926}\\
                                      &$\epsilon=0.02$ &3.476  &4.042   &19.782  &9.606   &5.868 &\revise{4.638}\\
                                      &$\epsilon=0.01$ &2.126  &3.106   &4.338   &9.430   &5.206 &\revise{1.356}\\
    \hline
    \multirow{4}{*}{CIFAR-10 (Test)} &$\epsilon=0.1$  &80.790 &69.760  &85.820  &64.880  &46.830 &\revise{67.940}\\
                                     &$\epsilon=0.05$ &17.990 &20.050  &71.240  &10.530  &14.700 &\revise{17.950}\\
                                     &$\epsilon=0.02$ &4.900  &4.470   &21.560  &2.490   &3.390 &\revise{5.220}\\
                                     &$\epsilon=0.01$ &1.870  &2.150   &4.440   &1.220   &1.430 &\revise{1.630}\\
    \hhline{|========|}
    \multirow{4}{*}{CIFAR-100 (Train)} &$\epsilon=0.1$  &84.848  &78.794  &96.980  &90.322  &83.720 &\revise{82.636}\\
                                       &$\epsilon=0.05$ &65.176  &42.456  &85.496  &64.512  &44.574 &\revise{47.410}\\
                                       &$\epsilon=0.02$ &22.348  &14.436  &34.382  &31.910  &21.004 &\revise{16.438}\\
                                       &$\epsilon=0.01$ &8.102   &7.708   &8.442   &28.640  &13.864 &\revise{2.664}\\
    \hline
    \multirow{4}{*}{CIFAR-100 (Test)} &$\epsilon=0.1$  &86.080  &78.640  &97.070  &88.380  &84.970 &\revise{83.040}\\
                                      &$\epsilon=0.05$ &67.960  &47.440  &86.980  &59.810  &49.160 &\revise{48.640}\\
                                      &$\epsilon=0.02$ &29.640  &20.750  &42.230  &16.190  &24.160 &\revise{19.740}\\
                                      &$\epsilon=0.01$ &10.110  &9.330   &16.070  &4.360   &9.470 &\revise{3.560}\\
    \hhline{|========|}
    \multirow{4}{*}{TinyImageNet (Train)} &$\epsilon=0.1$  &95.510  &95.261  &94.634  &96.780  &97.775 &\revise{98.279}\\
                                          &$\epsilon=0.05$ &73.192  &75.848  &72.553  &80.195  &90.642 &\revise{85.269}\\
                                          &$\epsilon=0.02$ &57.211  &57.009  &54.092  &66.979  &71.367 &\revise{43.215}\\
                                          &$\epsilon=0.01$ &53.432  &53.343  &47.967  &65.229  &66.845 &\revise{16.424}\\
    \hline
    \multirow{4}{*}{TinyImageNet (Test)} &$\epsilon=0.1$  &95.020  &94.700  &93.870  &95.030  &97.980 &\revise{98.530}\\
                                         &$\epsilon=0.05$ &62.350  &65.990  &64.280  &60.300  &87.700 &\revise{84.070}\\
                                         &$\epsilon=0.02$ &30.910  &31.000  &33.870  &21.480  &42.840 &\revise{42.570}\\
                                         &$\epsilon=0.01$ &18.200  &17.920  &19.150  &8.360  &17.120 &\revise{16.880}\\
    \hline
    \end{tabular}}
    \vspace{2mm}
    \caption{UAD-grad fooling rates with $\epsilon=0.1, 0.05, 0.02, 0.01 \cdot \mathbb{E}[\Vert \mathbf{X}\Vert_2]$.}
    \label{tab:fool_UADG}
\end{table}

\begin{table}[H]
    \renewcommand{\arraystretch}{1.1}
    \centering
    \footnotesize
    \resizebox{0.85\linewidth}{!}{
    \begin{tabular}{|c|c|c|c|c|c|c|c|}
    \cline{1-8}
    \hline
    \multicolumn{8}{|c|}{\textbf{UAP Fooling Rates}}  \\
    \hhline{|========|}
    \multicolumn{2}{|c|}{\backslashbox{\textbf{Dataset}}{\textbf{Model}}} & ResNet-18 & ResNet-34 & DenseNet-121 & CaffeNet & VGG-19 & \revise{EfficientNet}\\
         \hline
    \multirow{4}{*}{CIFAR-10 (Train)} &$\epsilon=0.1$  &56.288 &66.018  &80.908  &30.546  &29.944 &\revise{56.814}\\
                                      &$\epsilon=0.05$ &10.394 &10.916  &66.486  &11.812  &11.230 &\revise{5.188}\\
                                      &$\epsilon=0.02$ &3.018  &3.960   &7.104   &9.660   &5.366 &\revise{1.446}\\
                                      &$\epsilon=0.01$ &2.180  &3.102   &3.692   &9.500   &5.072 &\revise{0.662}\\
    \hline
    \multirow{4}{*}{CIFAR-10 (Test)} &$\epsilon=0.1$  &59.500 &67.180  &81.240  &31.160  &31.910 &\revise{56.640}\\
                                     &$\epsilon=0.05$ &12.960 &12.300  &68.040  &7.760   &11.030 &\revise{5.620}\\
                                     &$\epsilon=0.02$ &4.020  &4.200   &8.980   &2.330   &2.330 &\revise{1.740}\\
                                     &$\epsilon=0.01$ &1.520  &2.170   &3.960   &0.940   &1.150 &\revise{0.830}\\
    \hhline{|========|}
    \multirow{4}{*}{CIFAR-100 (Train)} &$\epsilon=0.1$  &81.466  &66.408  &80.606  &82.538  &62.346 &\revise{68.112}\\
                                       &$\epsilon=0.05$ &49.262  &31.618  &67.475  &59.060  &37.532 &\revise{29.208}\\
                                       &$\epsilon=0.02$ &11.462  &10.974  &22.131  &29.412  &15.318 &\revise{4.476}\\
                                       &$\epsilon=0.01$ &7.134   &7.208   &19.724   &28.376  &13.364 &\revise{1.938}\\
    \hline
    \multirow{4}{*}{CIFAR-100 (Test)} &$\epsilon=0.1$  &82.700 &67.850  &81.305  &80.710  &64.770 &\revise{69.960}\\
                                      &$\epsilon=0.05$ &54.460 &38.440  &68.878  &54.580  &42.370 &\revise{34.020}\\
                                      &$\epsilon=0.02$ &17.200 &16.640  &26.137  &9.270   &14.780 &\revise{5.880}\\
                                      &$\epsilon=0.01$ &8.060  &8.280   &11.253   &3.770   &6.400 &\revise{2.570}\\
    \hhline{|========|}
    \multirow{4}{*}{TinyImageNet (Train)} &$\epsilon=0.1$  &94.807  &95.835  &95.706  &93.622  &97.230 &\revise{97.853}\\
                                          &$\epsilon=0.05$ &73.221  &75.126  &70.545  &77.820  &87.035 &\revise{76.720}\\
                                          &$\epsilon=0.02$ &55.989  &55.817  &51.278  &66.454  &69.487 &\revise{31.901}\\
                                          &$\epsilon=0.01$ &52.773  &52.894  &46.956  &65.076  &66.724 &\revise{9.077}\\
    \hline
    \multirow{4}{*}{TinyImageNet (Test)} &$\epsilon=0.1$  &93.870  &95.250  &95.130  &89.770  &96.930 &\revise{98.270}\\
                                         &$\epsilon=0.05$ &61.300  &65.480  &61.530  &56.220  &82.040 &\revise{75.450}\\
                                         &$\epsilon=0.02$ &28.040  &27.750  &29.340  &19.060  &36.210 &\revise{30.800}\\
                                         &$\epsilon=0.01$ &16.860  &17.270  &16.880  &7.730   &16.100 &\revise{8.160}\\
    \hline
    \end{tabular}}
    \vspace{2mm}
    \caption{Gradient-based UAP fooling rates with $\epsilon=0.1, 0.05, 0.02, 0.01 \cdot \mathbb{E}[\Vert \mathbf{X}\Vert_2]$.}
    \label{tab:fool_UAP}
\end{table}

\begin{table}[h!]
\renewcommand{\arraystretch}{1.02}
\begin{subtable}[t]{0.49\linewidth}
    \centering
    \footnotesize
    \begin{tabular}{|l|l|l|l|l|l|l|}
    \hline
    \multicolumn{7}{|c|}{\revise{\textbf{UAD Cosine Similarities for CIFAR-10}}}  \\
    \hhline{|=======|}
    {\footnotesize \revise{Tar / Src}} & \revise{R18} & \revise{R34} & \revise{D121} & \revise{Caffe} & \revise{VGG} & \revise{Eff2s}\\
    \hline
    \revise{R18} & \revise{\textbf{1.00}} & \revise{-0.70} & \revise{0.54} & \revise{-0.43} & \revise{-0.36} & \revise{-0.16} \\
    \hline
    \revise{R34} & \revise{-0.70} & \revise{\textbf{1.00}} & \revise{0.73} & \revise{0.30} & \revise{-0.50} & \revise{-0.41}  \\
    \hline
    \revise{D121} & \revise{0.54} & \revise{0.73} & \revise{\textbf{1.00}} & \revise{0.31} & \revise{-0.20} & \revise{0.86} \\
    \hline
    \revise{Caffe} & \revise{-0.43} & \revise{0.30} & \revise{0.31} & \revise{\textbf{1.00}} & \revise{-0.26} & \revise{0.60} \\
    \hline
    \revise{VGG} & \revise{-0.36} & \revise{-0.50} & \revise{-0.20} & \revise{-0.26} & \revise{\textbf{1.00}} &\revise{-0.34} \\
    \hline
    \revise{Eff2s} & \revise{-0.16} & \revise{-0.41} &\revise{0.86} &\revise{0.60} &\revise{-0.34} &\revise{\textbf{1.00}}\\
    \hhline{|=======|}
    \hline
    \multicolumn{7}{|c|}{\revise{\textbf{UAP Cosine Similarities for CIFAR-10}}}  \\
    \hhline{|=======|}
    {\footnotesize \revise{Tar / Src}} & \revise{R18} & \revise{R34} & \revise{D121} & \revise{Caffe} & \revise{VGG} & \revise{Eff2s}\\
    \hline
    \revise{R18} & \revise{\textbf{1.00}} & \revise{-0.04} & \revise{-0.00} & \revise{-0.02} & \revise{0.04} & \revise{-0.06} \\
    \hline
    \revise{R34} & \revise{-0.04} & \revise{\textbf{1.00}} & \revise{0.03} & \revise{0.01} & \revise{-0.01} & \revise{0.00}  \\
    \hline
    \revise{D121} & \revise{-0.00} & \revise{0.03} & \revise{\textbf{1.00}} & \revise{0.01} & \revise{-0.00} & \revise{0.03} \\
    \hline
    \revise{Caffe} & \revise{-0.03} & \revise{-0.01} & \revise{-0.01} & \revise{\textbf{1.00}} & \revise{-0.03} & \revise{-0.02} \\
    \hline
    \revise{VGG} & \revise{0.04} & \revise{-0.01} & \revise{-0.00} & \revise{-0.03} & \revise{\textbf{1.00}} &\revise{0.06} \\
    \hline
    \revise{Eff2s} & \revise{-0.06} & \revise{0.00} &\revise{0.03} &\revise{-0.02} &\revise{0.06} &\revise{\textbf{1.00}}\\
    \hhline{|=======|}
    \end{tabular}
    \caption{Cosine similarity scores for UAD \& UAP on CIFAR-10.}
    \label{tab:cosine_c10}
    \end{subtable}\hfill
    \begin{subtable}[t]{.49\linewidth}
        \centering
        \footnotesize
    \begin{tabular}{|c|c|c|c|c|c|c|}
    \hline
    \multicolumn{7}{|c|}{\textbf{UAD TFR for CIFAR-10}}  \\
    \hhline{|=======|}
    {\footnotesize \revise{Tar / Src}} & R18 & R34 & D121 & Caffe & VGG &Eff2s\\
    \hline
    R18 & \textbf{0.884} & 0.728 & 0.594 & 0.417 & 0.550 &0.597\\
    \hline
    R34 & 0.772 & \textbf{0.854} & 0.532 & 0.406 & 0.503 & 0.456\\
    \hline
    D121 & 0.788 & 0.723 & \textbf{0.869} & 0.411 & 0.547 & 0.488\\
    \hline
    Caffe & 0.324 & 0.320 & 0.267 & \textbf{0.740} & 0.447 & 0.344\\
    \hline
    VGG & 0.295 & 0.403 & 0.145 & 0.241 & \textbf{0.712} & 0.394\\
    \hline
    Eff2s &0.321 & 0.341 & 0.294 & 0.231 & 0.280 & \textbf{0.694}\\
    \hhline{|=======|}
    \hline
    \multicolumn{7}{|c|}{\textbf{UAP TFR for CIFAR-10}}  \\
    \hhline{|=======|}
    {\footnotesize \revise{Tar / Src}} & R18 & R34 & D121 & Caffe & VGG &Eff2s\\
    \hline
    R18 & \textbf{0.595} & 0.610 & 0.378 & 0.298 & 0.305 &0.465\\
    \hline
    R34 & 0.483 & \textbf{0.672} & 0.366 & 0.282 & 0.325 & 0.450\\
    \hline
    D121 & 0.481 & 0.549 & \textbf{0.812} & 0.333 & 0.316 & 0.441\\
    \hline
    Caffe & 0.257 & 0.281 & 0.232 & \textbf{0.312} & 0.273 & 0.313\\
    \hline
    VGG & 0.199 & 0.230 & 0.111 & 0.183 & \textbf{0.319} & 0.347\\
    \hline
    Eff2s &0.105 & 0.130 & 0.109 & 0.094 & 0.148 & \textbf{0.566}\\
    \hhline{|=======|}
    \end{tabular}   
    \caption{Transferred fooling rates for UAD \& UAP on CIFAR-10.}
    \label{tab:tfr_c10}
    \end{subtable}
    \caption{UAD and UAP cross-network transferability comparison.}
\end{table}

\begin{table}[h!]
\renewcommand{\arraystretch}{1.02}
\begin{subtable}[t]{0.49\linewidth}
    \centering
    \footnotesize
    \begin{tabular}{|l|l|l|l|l|l|l|}
    \hline
    \multicolumn{7}{|c|}{\revise{\textbf{UAD Cosine Similarities for CIFAR-100}}}  \\
    \hhline{|=======|}
    {\footnotesize \revise{Tar / Src}} & \revise{R18} & \revise{R34} & \revise{D121} & \revise{Caffe} & \revise{VGG} & \revise{Eff2s}\\
    \hline
    \revise{R18} & \revise{\textbf{1.00}} & \revise{0.41} & \revise{-0.53} & \revise{-0.50} & \revise{-0.49} & \revise{-0.23} \\
    \hline
    \revise{R34} & \revise{0.41} & \revise{\textbf{1.00}} & \revise{0.38} & \revise{-0.53} & \revise{0.41} & \revise{-0.23}  \\
    \hline
    \revise{D121} & \revise{-0.53} & \revise{0.38} & \revise{\textbf{1.00}} & \revise{-0.50} & \revise{-0.47} & \revise{-0.13} \\
    \hline
    \revise{Caffe} & \revise{-0.50} & \revise{-0.53} & \revise{-0.50} & \revise{\textbf{1.00}} & \revise{0.39} & \revise{-0.37} \\
    \hline
    \revise{VGG} & \revise{-0.49} & \revise{0.41} & \revise{-0.47} & \revise{0.39} & \revise{\textbf{1.00}} &\revise{-0.47} \\
    \hline
    \revise{Eff2s} & \revise{-0.23} & \revise{-0.23} &\revise{-0.13} &\revise{-0.37} &\revise{-0.47} &\revise{\textbf{1.00}}\\
    \hhline{|=======|}
    \hline
    \multicolumn{7}{|c|}{\revise{\textbf{UAP Cosine Similarities for CIFAR-100}}}  \\
    \hhline{|=======|}
    {\footnotesize \revise{Tar / Src}} & \revise{R18} & \revise{R34} & \revise{D121} & \revise{Caffe} & \revise{VGG} & \revise{Eff2s}\\
    \hline
    \revise{R18} & \revise{\textbf{1.00}} & \revise{0.02} & \revise{-0.07} & \revise{0.04} & \revise{0.03} & \revise{-0.07} \\
    \hline
    \revise{R34} & \revise{0.02} & \revise{\textbf{1.00}} & \revise{-0.01} & \revise{-0.06} & \revise{-0.02} & \revise{0.02}  \\
    \hline
    \revise{D121} & \revise{-0.07} & \revise{-0.01} & \revise{\textbf{1.00}} & \revise{0.02} & \revise{-0.03} & \revise{-0.02} \\
    \hline
    \revise{Caffe} & \revise{0.04} & \revise{-0.06} & \revise{0.02} & \revise{\textbf{1.00}} & \revise{0.01} & \revise{0.01} \\
    \hline
    \revise{VGG} & \revise{0.03} & \revise{-0.02} & \revise{-0.03} & \revise{0.01} & \revise{\textbf{1.00}} &\revise{-0.05} \\
    \hline
    \revise{Eff2s} & \revise{-0.07} & \revise{0.02} &\revise{-0.02} &\revise{0.01} &\revise{-0.05} &\revise{\textbf{1.00}}\\
    \hhline{|=======|}
    \end{tabular}
    \caption{Cosine similarity scores for UAD \& UAP on CIFAR-100.}
    \label{tab:cosine_c100}
    \end{subtable}\hfill
    \begin{subtable}[t]{.49\linewidth}
        \centering
        \footnotesize
    \begin{tabular}{|c|c|c|c|c|c|c|}
    \hline
    \multicolumn{7}{|c|}{\textbf{UAD TFR for CIFAR-100}}  \\
    \hhline{|=======|}
    {\footnotesize \revise{Tar / Src}} & R18 & R34 & D121 & Caffe & VGG &Eff2s\\
    \hline
    R18 & \textbf{0.919} & 0.770 & 0.863 & 0.699 & 0.708 &0.660\\
    \hline
    R34 & 0.767 & \textbf{0.893} & 0.783 & 0.598 & 0.732 & 0.578\\
    \hline
    D121 & 0.794 & 0.763 & \textbf{0.975} & 0.743 & 0.773 & 0.660\\
    \hline
    Caffe & 0.632 & 0.564 & 0.688 & \textbf{0.953} & 0.617 & 0.547\\
    \hline
    VGG & 0.673 & 0.642 & 0.747 & 0.746 & \textbf{0.870} & 0.608\\
    \hline
    Eff2s &0.561 & 0.520 & 0.612 & 0.678 & 0.693 & \textbf{0.848}\\
    \hhline{|=======|}
    \hline
    \multicolumn{7}{|c|}{\textbf{UAP TFR for CIFAR-100}}  \\
    \hhline{|=======|}
    {\footnotesize \revise{Tar / Src}} & R18 & R34 & D121 & Caffe & VGG &Eff2s\\
    \hline
    R18 & \textbf{0.827} & 0.724 & 0.869 & 0.726 & 0.688 &0.594\\
    \hline
    R34 & 0.596 & \textbf{0.679} & 0.797 & 0.616 & 0.620 & 0.562\\
    \hline
    D121 & 0.886 & 0.713 & \textbf{0.813} & 0.755 & 0.653 & 0.566\\
    \hline
    Caffe & 0.624 & 0.576 & 0.701 & \textbf{0.807} & 0.592 & 0.505\\
    \hline
    VGG & 0.392 & 0.636 & 0.773 & 0.692 & \textbf{0.648} & 0.616\\
    \hline
    Eff2s &0.303 & 0.365 & 0.256 & 0.351 & 0.389 & \textbf{0.700}\\
    \hhline{|=======|}
    \end{tabular}   
    \caption{Transferred fooling rates for UAD \& UAP on CIFAR-100.}
    \label{tab:tfr_c100}
    \end{subtable}
    \caption{UAD and UAP cross-network transferability comparison.}
\end{table}

\begin{table}[h!]
\renewcommand{\arraystretch}{1.02}
\begin{subtable}[t]{0.49\linewidth}
    \centering
    \footnotesize
    \begin{tabular}{|l|l|l|l|l|l|l|}
    \hline
    \multicolumn{7}{|c|}{\revise{\textbf{UAD Cosine Similarities for TinyImageNet}}}  \\
    \hhline{|=======|}
    {\footnotesize \revise{Tar / Src}} & \revise{R18} & \revise{R34} & \revise{D121} & \revise{Caffe} & \revise{VGG} & \revise{Eff2s}\\
    \hline
    \revise{R18} & \revise{\textbf{1.00}} & \revise{0.59} & \revise{0.27} & \revise{0.29} & \revise{-0.64} & \revise{-0.52} \\
    \hline
    \revise{R34} & \revise{0.59} & \revise{\textbf{1.00}} & \revise{-0.65} & \revise{-0.39} & \revise{-0.54} & \revise{0.58}  \\
    \hline
    \revise{D121} & \revise{0.27} & \revise{-0.65} & \revise{\textbf{1.00}} & \revise{0.34} & \revise{-0.45} & \revise{0.79} \\
    \hline
    \revise{Caffe} & \revise{0.29} & \revise{-0.39} & \revise{0.34} & \revise{\textbf{1.00}} & \revise{-0.37} & \revise{-0.17} \\
    \hline
    \revise{VGG} & \revise{-0.64} & \revise{-0.54} & \revise{-0.45} & \revise{-0.37} & \revise{\textbf{1.00}} &\revise{-0.29} \\
    \hline
    \revise{Eff2s} & \revise{-0.52} & \revise{0.58} &\revise{0.79} &\revise{-0.17} &\revise{-0.29} &\revise{\textbf{1.00}}\\
    \hhline{|=======|}
    \hline
    \multicolumn{7}{|c|}{\revise{\textbf{UAP Cosine Similarities for TinyImageNet}}}  \\
    \hhline{|=======|}
    {\footnotesize \revise{Tar / Src}} & \revise{R18} & \revise{R34} & \revise{D121} & \revise{Caffe} & \revise{VGG} & \revise{Eff2s}\\
    \hline
    \revise{R18} & \revise{\textbf{1.00}} & \revise{0.02} & \revise{-0.03} & \revise{-0.04} & \revise{0.02} & \revise{0.05} \\
    \hline
    \revise{R34} & \revise{0.02} & \revise{\textbf{1.00}} & \revise{0.01} & \revise{0.01} & \revise{0.04} & \revise{-0.01}  \\
    \hline
    \revise{D121} & \revise{-0.03} & \revise{0.01} & \revise{\textbf{1.00}} & \revise{-0.00} & \revise{-0.02} & \revise{0.01} \\
    \hline
    \revise{Caffe} & \revise{-0.04} & \revise{0.01} & \revise{-0.00} & \revise{\textbf{1.00}} & \revise{-0.00} & \revise{0.00} \\
    \hline
    \revise{VGG} & \revise{0.02} & \revise{0.04} & \revise{-0.02} & \revise{-0.00} & \revise{\textbf{1.00}} &\revise{-0.03} \\
    \hline
    \revise{Eff2s} & \revise{0.05} & \revise{-0.01} &\revise{0.01} &\revise{0.00} &\revise{-0.03} &\revise{\textbf{1.00}}\\
    \hhline{|=======|}
    \end{tabular}
    \caption{Cosine similarity scores for UAD \& UAP on TinyImageNet.}
    \label{tab:cosine_tim}
    \end{subtable}\hfill
    \begin{subtable}[t]{.49\linewidth}
        \centering
        \footnotesize
    \begin{tabular}{|c|c|c|c|c|c|c|}
    \hline
    \multicolumn{7}{|c|}{\textbf{UAD TFR for TinyImageNet}}  \\
    \hhline{|=======|}
    {\footnotesize \revise{Tar / Src}} & R18 & R34 & D121 & Caffe & VGG &Eff2s\\
    \hline
    R18 & \textbf{0.982} & 0.956 & 0.943 & 0.720 & 0.887 &0.836\\
    \hline
    R34 & 0.938 & \textbf{0.986} & 0.945 & 0.668 & 0.878 & 0.826\\
    \hline
    D121 & 0.926 & 0.914 & \textbf{0.983} & 0.699 & 0.836 & 0.901\\
    \hline
    Caffe & 0.695 & 0.743 & 0.760 & \textbf{0.981} & 0.787 & 0.883\\
    \hline
    VGG & 0.874 & 0.886 & 0.826 & 0.723 & \textbf{0.994} & 0.853\\
    \hline
    Eff2s &0.866 & 0.915 & 0.912 & 0.874 & 0.877 & \textbf{0.974}\\
    \hhline{|=======|}
    \hline
    \multicolumn{7}{|c|}{\textbf{UAP TFR for TinyImageNet}}  \\
    \hhline{|=======|}
    {\footnotesize \revise{Tar / Src}} & R18 & R34 & D121 & Caffe & VGG &Eff2s\\
    \hline
    R18 & \textbf{0.939} & 0.925 & 0.932 & 0.726 & 0.901 &0.853\\
    \hline
    R34 & 0.921 & \textbf{0.952} & 0.903 & 0.690 & 0.880 & 0.832\\
    \hline
    D121 & 0.892 & 0.908 & \textbf{0.951} & 0.703 & 0.852 & 0.768\\
    \hline
    Caffe & 0.808 & 0.802 & 0.769 & \textbf{0.898} & 0.811 & 0.818\\
    \hline
    VGG & 0.931 & 0.940 & 0.909 & 0.835 & \textbf{0.969} & 0.801\\
    \hline
    Eff2s &0.890 & 0.890 & 0.831 & 0.720 & 0.885 & \textbf{0.983}\\
    \hhline{|=======|}
    \end{tabular}   
    \caption{Transferred fooling rates for UAD \& UAP on TinyImageNet.}
    \label{tab:tfr_tim}
    \end{subtable}
    \caption{UAD and UAP cross-network transferability comparison.}
\end{table} 
\end{appendices}

\end{document}